\RequirePackage[svgnames]{xcolor}

\documentclass[11pt,logo,letterpaper]{berkeley}

\usepackage[utf8]{inputenc} %
\usepackage[T1]{fontenc}    %
\usepackage{lmodern}
\usepackage{textcomp}
\usepackage[svgnames]{xcolor}
\usepackage{asymptote}
\usepackage[comma,authoryear,compress]{natbib}
\usepackage{hyperref}
\usepackage{algorithm}
\usepackage{algorithmicx}
\usepackage{algpseudocode}
\usepackage{microtype}
\usepackage{graphicx}
\expandafter\def\csname ver@subfig.sty\endcsname{}
\usepackage{booktabs} %
\usepackage{float}
\usepackage{bigstrut}
\usepackage{tikz}
\usetikzlibrary{tikzmark}
\makeatletter
\newcommand*\myfontsize{%
  \@setfontsize\myfontsize{7}{8}%
}
\makeatother
\definecolor{myred}{rgb}{0.7, 0.3, 0.0}
\definecolor{myblue}{HTML}{054488}
\definecolor{mygreen}{HTML}{056b34}

\newcolumntype{R}[1]{>{\raggedleft\let\newline\\\arraybackslash\hspace{0pt}}m{#1}}

\usetikzlibrary{intersections}

\definecolor{darkgreen}{rgb}{0.0, 0.42, 0.24}

\usepackage{amsthm}
\usepackage{nicefrac}
\usepackage{subcaption}
\usepackage{caption}
\usepackage{graphicx}
\usepackage{cleveref}
\usepackage{bxcoloremoji}
\usepackage{listings}
\usepackage{float}
\usepackage{longtable}
\lstdefinestyle{python}{
    language=Python,
    basicstyle=\ttfamily\footnotesize,
    keywordstyle=\color{blue}\bfseries,
    commentstyle=\color{green},
    stringstyle=\color{red},
    numberstyle=\tiny\color{gray},
    showstringspaces=false,
    frame=single,
    breaklines=true,
    backgroundcolor=\color{lightgray!20}
}

\usepackage{hyperref}       %
\usepackage{url}            %
\usepackage{booktabs}       %
\usepackage{nicefrac}       %
\usepackage{microtype}      %
\usepackage{graphicx}
\usepackage{subcaption} % Ensure subfigure or subtable environments are used if this package is included.
\usepackage{wrapfig}
\usepackage{lipsum}
\usepackage{enumitem}
\usepackage{stackengine}
\usepackage[font=small,labelfont=bf]{caption}
\usepackage{color}
\usepackage{adjustbox}
\usepackage{tikz}
\usepackage{tcolorbox}
\usepackage{rotating}
\usepackage{makecell}
\usepackage{colortbl}    % \rowcolor
\usepackage{multirow}    % 多行合并（如果以后需要）
\usepackage{pifont}      % \ding 对勾/叉
\usepackage{xcolor}      % 自定义颜色
\usepackage{array}       % 控制列宽（可选）
\usepackage{amsmath}
\usepackage{amsfonts}       %
\usepackage{xspace}
\usepackage{mathtools}
\definecolor{oursblue}{RGB}{230,240,255} % 淡蓝行底色

\newcommand{\x}{\mathbf{x}}

\definecolor{blanchedalmond}{rgb}{1.0, 0.92, 0.8}
\definecolor{carmine}{rgb}{0.59, 0.0, 0.09}
\definecolor{lightblue}{rgb}{0.22,0.45,0.70}%

\renewcommand{\mathbf}{\boldsymbol}

\makeatletter
\def\Ddots{\mathinner{\mkern1mu\raise\p@
\vbox{\kern7\p@\hbox{.}}\mkern2mu
\raise4\p@\hbox{.}\mkern2mu\raise7\p@\hbox{.}\mkern1mu}}
\makeatother

\definecolor{amaranth}{rgb}{0.9, 0.17, 0.31}
\definecolor{antiquebrass}{rgb}{0.8, 0.58, 0.46}
\definecolor{antiquefuchsia}{rgb}{0.57, 0.36, 0.51}
\definecolor{chromeyellow}{rgb}{0.31, 0.47, 0.26}

\newtcolorbox{AIbox}[2][]{aibox,title=#2,#1}
\definecolor{lightblue}{rgb}{0.22,0.45,0.70}%
\definecolor{Gray}{gray}{0.95}
\definecolor{Cornsilk}{rgb}{1.0, 0.97, 0.86}
\usepackage{enumitem}

\makeatother

\definecolor{myred}{rgb}{0.7, 0.3, 0.0}
\definecolor{myblue}{HTML}{054488}
\definecolor{mygreen}{HTML}{056b34}
\definecolor{myorange}{HTML}{ff8800}
\definecolor{mypurple}{HTML}{8400ff}
\definecolor{mypink}{HTML}{f7acb9}

\definecolor{myred}{rgb}{0.7, 0.3, 0.0}
\definecolor{myblue}{HTML}{054488}
\definecolor{mygreen}{HTML}{056b34}
\definecolor{tiktokpink}{HTML}{E91E63}
\definecolor{tiktokpurple}{HTML}{673AB7}
\definecolor{tiktokgray}{HTML}{9E9E9E}

\newcommand{\mytitle}{Digital Twin-Driven Pavement Health Monitoring and Maintenance Optimization Using Graph Neural Networks}

\usepackage{amssymb}

\title{\mytitle}

\runningtitle{Graph-based Digital Twins for Pavement Health Monitoring and Maintenance}

\author{
  Mohsin Mahmud Topu$^{1,2}$
  Mahfuz Ahmed Anik$^{1,2}$\\
  Azmine Toushik Wasi$^{1,2}$
  Md Manjurul Ahsan$^{1,3}$
}

\affil{$^1$Computational Intelligence and Operations Laboratory (CIOL), Bangladesh\\
\vspace{-2.5mm}
$^2$Shahjalal University of Science and Technology, Sylhet, Bangladesh\\
\vspace{-2.5mm}
$^3$University of Oklahoma, Oval Norman, OK 73019, USA\\
}

\begin{document}

\begin{abstract}
\textbf{Abstract:}
Pavement infrastructure monitoring is challenged by complex spatial dependencies, changing environmental conditions, and non-linear deterioration across road networks. Traditional Pavement Management Systems (PMS) remain largely reactive, lacking real-time intelligence for failure prevention and optimal maintenance planning. To address this, we propose a unified Digital Twin (DT) and Graph Neural Network (GNN) framework for scalable, data-driven pavement health monitoring and predictive maintenance. Pavement segments and spatial relations are modeled as graph nodes and edges, while real-time UAV, sensor, and LiDAR data stream into the DT. The inductive GNN learns deterioration patterns from graph-structured inputs to forecast distress and enable proactive interventions. Trained on a real-world-inspired dataset with segment attributes and dynamic connectivity, our model achieves an R2 of 0.3798, outperforming baseline regressors and effectively capturing non-linear degradation. We also develop an interactive dashboard and reinforcement learning module for simulation, visualization, and adaptive maintenance planning. This DT-GNN integration enhances forecasting precision and establishes a closed feedback loop for continuous improvement, positioning the approach as a foundation for proactive, intelligent, and sustainable pavement management, with future extensions toward real-world deployment, multi-agent coordination, and smart-city integration.

\vspace{0.5cm}

\coloremojicode{1F4C5} \textbf{Date}: Sep 2, 2025

\coloremojicode{1F4E7} \textbf{Correspondence}: Mohsin Mahmud Topu~(\href{mailto:mohsin69@student.sust.edu}{mohsin69@student.sust.edu})

% \coloremojicode{1F4BB} \textbf{Code}: \href{}{Github Repo} \qquad
% \coloremojicode{1F9E0} \textbf{Model}: \href{}{HuggingFace}

\end{abstract}

\maketitle

\section{Introduction}
Pavement infrastructure forms the backbone of modern transportation systems, enabling the movement of goods and people and thus supporting economic activity and social connectivity \citep{khan2023impact}. The existing pavement infrastructure faces several challenges, including adapting to the growing demands and priorities of maintaining and improving service, as well as expanding its lifespan. Pavement assets are the main contributors to energy consumption and emissions \citep{mantalovas2020european}. Managing this is critical for the safe and efficient movement of users and the economy, and it must be managed efficiently \citep{liu2022integrating}. Its operation and maintenance phase is usually a point of concern due to the cruciality of any possible improvement \citep{lu2019generating}. Deteriorating pavements contribute to prolonged travel durations, increased fuel consumption, increased vehicle operating costs, and a higher risk of traffic accidents. Therefore, proactive maintenance and robust monitoring of pavement health are imperative to secure the longevity and optimal functionality of road networks. The structural integrity of pavements must be ensured from the beginning of the construction phase, as they are subjected to large vehicular loads \citep{de2009guide}. Traditional Pavement Management Systems (PMS) predominantly employ a reactive maintenance strategy, where interventions are initiated only upon the appearance of pavement failures \citep{tamagusko2024machine}. However, this approach proves to be less cost-effective compared to a proactive strategy, which aims to prevent deterioration before failure occurs \citep{talaghat2024digital}. The inefficacy of traditional PMS is attributed to financial constraints and scheduling limitations associated with road condition monitoring, resulting in ineffective maintenance practices \citep{talaghat2024digital}. Pavements are conventionally designed for a lifespan of approximately 40 years, with maintenance interventions scheduled at 10-year intervals \citep{sierra2022development}. But these predetermined schedules may not align with actual deterioration trajectories \citep{garcia2022incorporating}. Such misalignment results in either premature structural failures or excessive maintenance costs \citep{fernando2020markovian}. Therefore, PMS should shift towards a proactive maintenance paradigm, necessitating continuous pavement monitoring, systematic data acquisition, and advanced analytical frameworks leveraging digital technologies and innovative computational tools.

Recent advancements in digital technologies have unveiled revolutionary opportunities for pavement maintenance and management \citep{kodikara2024reimagining}. Notably, sophisticated methodologies such as DTs and neural networks are groundbreaking conventional paradigms in asset monitoring and predictive maintenance \citep{borovkov2024synergistic}. A DT constitutes a highly detailed, real-time virtual representation of a physical asset, continuously updated through the integration of heterogeneous data sources, including sensor-derived information, computational simulations, and historical archives \citep{singh2021digital}. \cite{glaessgen2012digital} define DTs as advanced cyber-physical systems that encapsulate the entire lifecycle of their corresponding physical counterparts. Over the last decade, the rapid evolution of digital technologies has significantly enhanced the applicability of DTs across diverse domains, including manufacturing, agriculture, healthcare, and the development of intelligent and sustainable urban environments \citep{fuller2020digital}. The primary objectives of DT implementations are to enhance system efficiency and optimize performance through real-time monitoring, predictive maintenance strategies, and data-driven decision-making frameworks \citep{ali2024enabling}. In the domain of pavement management, DTs offer a dynamic and continuously evolving representation of road conditions by integrating real-time sensor data with historical deterioration patterns \citep{bertolini2024semantic}. A pertinent example is the use of a DT-enabled system for asphalt pavements, which leverages thermal expansion and contraction analyses to predict surface crack formations induced by climatic fluctuations \citep{barisic2021thermal}. This predictive capability facilitates preemptive interventions, thereby mitigating substantial structural degradation and prolonging pavement service life \citep{zakharchenko2023digital}. 

Despite the crucial insights facilitated by DT technology, the precision and efficacy of its decision-making and predictive functionalities can be appreciably augmented through the integration of advanced analytical methodologies \citep{wettewa2024graph}. One such approach is the incorporation of GNNs, which significantly enhance the capacity to elucidate intricate spatial and temporal datasets, thereby enabling a more robust evaluation of pavement health and deterioration trends \citep{nippani2023graph,wasi2025graphneuralnetworkssupply}. GNNs depict a sophisticated class of machine learning models specifically designed to process data structured in the form of graphs or networks, making them particularly suitable for applications that require the analysis of interconnected systems \citep{khemani2024review}. More specifically, GNNs are engineered to handle graph-based data representations that inherently capture the spatial relationships among various infrastructure components, such as distinct pavement segments \citep{feng2021gcn}. This capability facilitates a more comprehensive understanding of the interdependencies between multiple variables and their collective impact on overall infrastructure integrity \citep{khemani2024review}. While these innovations hold immense transformative potential, their application in pavement management remains relatively nascent \citep{chamorro2024prediction}. Although DT technology has been widely deployed in multiple sectors, including architecture, logistics, and manufacturing, its adoption within pavement infrastructure management remains markedly underexplored \citep{oditallah2025review}. Similarly, while GNNs have displayed exceptional efficacy in processing complex spatial data, their full potential in the domain of pavement health monitoring remains largely untapped \citep{gao2024considering}. This gap, coupled with the inherent limitations of traditional pavement management systems, emphasizes the importance of a more integrated, data-driven, and anticipatory approach to pavement infrastructure maintenance. The implementation of predictive models and real-time feedback mechanisms can substantially improve precision, reduce decision-making uncertainties, and enhance system reliability, thereby fostering a paradigm shift toward more precise, resilient, and cost-effective pavement health monitoring and maintenance strategies \citep{tong2025stgan}.

\begin{figure}
    \centering
    \includegraphics[width=\linewidth]{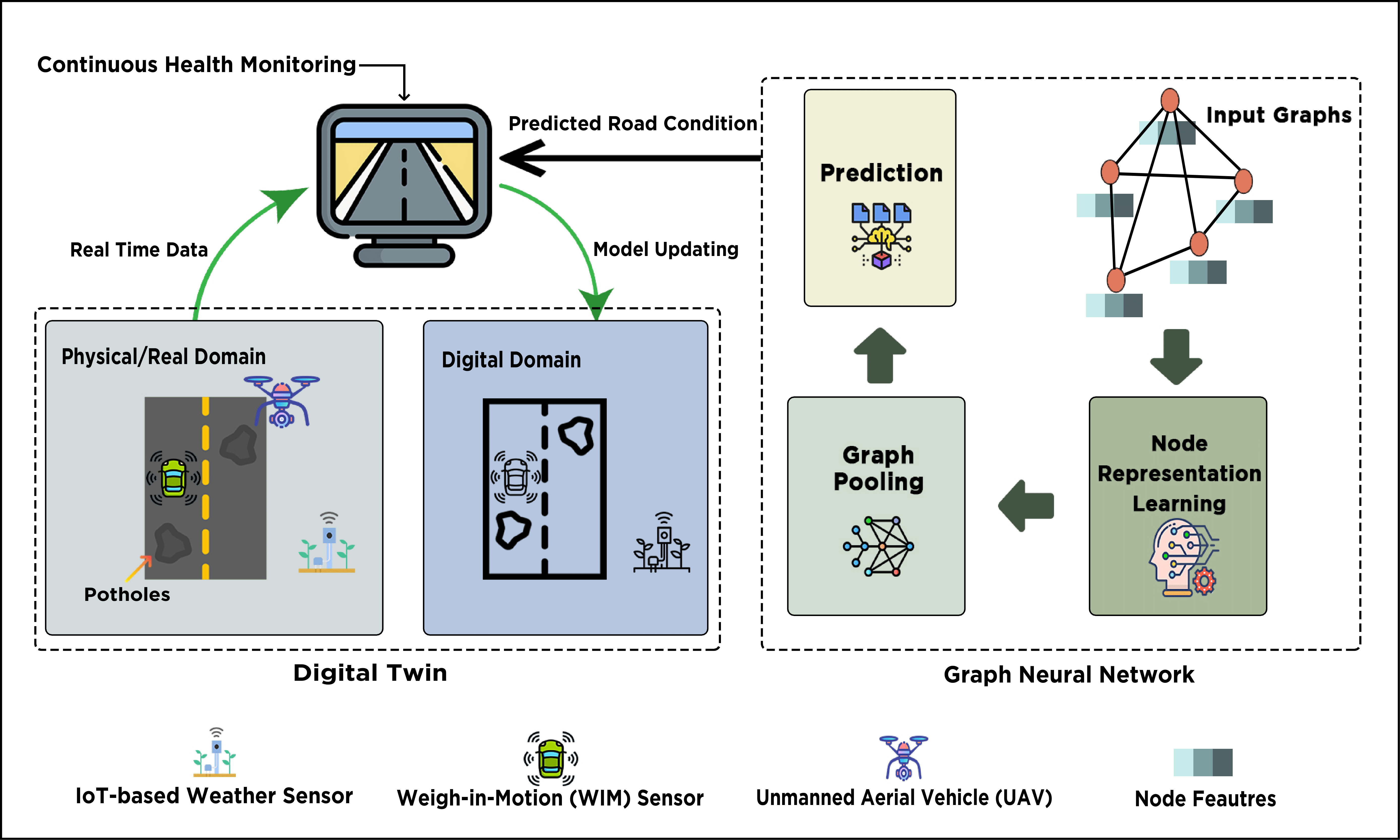}
    \caption{Theoretical GNN-Based DT Framework for Pavement Health Monitoring }
    \label{fig:Main1}
\end{figure}

To overcome the limitations of reactive pavement maintenance and effectively address the complex spatiotemporal nature of pavement deterioration, we introduce a GNN-enhanced DT framework for real-time condition monitoring and optimized intervention planning. The proposed system integrates a graph neural network that models spatial dependencies between pavement segments with a dynamic DT platform continuously updated using live sensor data, historical condition records, and UAV-based assessments. The GNN is trained on features including segment length, material type, traffic volume, and age, and employs a message-passing mechanism tailored to mitigate overfitting in sparse or noisy datasets, thereby enabling accurate modeling of both localized distress and system-wide degradation trends \citep{mukhopadhyay2024sparsity}. Simultaneously, the DT supports what-if scenario analysis, allowing practitioners to simulate diverse maintenance strategies under varying environmental and traffic conditions before field deployment \citep{hodavand2023digital}. This closed feedback loop—where updated observations refine GNN predictions, and predictive insights inform DT simulations—enhances both forecasting precision and operational decision-making. Figure~\ref{fig:Main1} illustrates this framework, highlighting the bidirectional interaction between DT and GNN modules, and depicting the full data flow from collection and integration to prediction and decision support. Empirical evaluation using a real-world-inspired dataset demonstrates the superior predictive performance of the proposed GNN, achieving the highest R² score (0.3798), with balanced MAE (31.34) and RMSE (38.93), outperforming conventional regression models. Furthermore, the integrated system facilitates cost-benefit analysis, contingency planning, and resource optimization, supporting data-driven decisions that minimize disruptions and extend pavement lifespan \citep{li2024digital}. By combining predictive learning with real-time simulation, this DT-GNN framework presents a scalable, intelligent, and sustainable approach to urban pavement infrastructure management, laying the groundwork for future enhancements such as reinforcement learning-based scheduling and integration within smart city ecosystems.

Recognizing the critical role of pavements in modern infrastructure and the need for lifecycle-optimized maintenance, this research offers the following key contributions:

\begin{enumerate}
    \item We propose an integrated Digital Twin (DT) framework that continuously synchronizes real-time data from UAVs, LiDAR scans, embedded sensors, and historical pavement records to provide dynamic, high-fidelity monitoring of road infrastructure.
    
    \item A graph-based predictive model is developed using Graph Neural Networks (GNNs), capturing spatiotemporal dependencies among pavement segments and learning from physical attributes, traffic loads, and environmental conditions to forecast deterioration trends with high accuracy.

    \item The DT-GNN integration enables interactive simulation and what-if scenario analysis, allowing practitioners to evaluate maintenance strategies under varying conditions and optimize intervention schedules in a virtual environment before field deployment.

    \item A comprehensive comparative evaluation demonstrates the proposed system’s superior predictive performance and generalization capacity over traditional machine learning models, highlighting its scalability, cost-efficiency, and potential for proactive pavement lifecycle management.
\end{enumerate}

In the following sections, we systematically elaborate on the core components and contributions of our proposed framework. Section \ref{sec:Background} introduces the foundational concepts of DTs and GNNs, emphasizing their relevance to pavement health monitoring. In Section \ref{sec:RelatedWorks}, we review related work on DT-enabled infrastructure applications and GNN-based predictive modeling, identifying critical research gaps. Section \ref{sec:motivation} articulates the rationale for integrating DT and GNN within a unified framework. Section \ref{sec: framework-emthodology} presents the proposed architecture, detailing the data acquisition pipeline, graph construction strategy, and simulation methodology. Section \ref{sec: experiments} evaluates the framework through experiments and case studies under diverse infrastructure scenarios. Section \ref{sec:discussion} discusses the broader implications of our approach, focusing on predictive performance, operational efficiency, applications and scalability. Finally, Section \ref{sec:conclusion} summarizes the key contributions and reflects on the potential impact of DT-GNN integration in advancing data-driven infrastructure management.

\section{Background and Fundamentals} \label{sec:Background}
In this section, we present the foundational concepts behind DTs and GNNs, focusing on their relevance to pavement health monitoring. We explore how DTs enable real-time simulation and decision-making, while GNNs model complex spatial relationships to predict deterioration. Together, these technologies offer a powerful framework for proactive and data-driven infrastructure maintenance.

\subsection{Digital Twin} 

Industries and academies define a DT in many ways \citep{trauer2020digital}. For instance, according to some, a DT is a high-fidelity virtual representation of a physical object, system, or process, a continuously updated model that interacts with the physical system throughout its life cycle \citep{grieves2017digital}. Other widely used definitions regard the need to exchange information between the two spaces involving sensors, and models with real-time data to reflect its actual operational state \citep{negri2017review}. This dynamic model facilitates simulation, monitoring, and predictive analysis. Leveraging machine learning and reasoning to optimize decision-making it is regarded as a new paradigm in simulation \citep{hao2024catalyzing}. The true strength of a DT lies in its ability to provide a near-real-time comprehensive linkage between the physical and virtual systems \citep{semeraro2021digital, anik5212671digital}. Its core principle is the bidirectional flow of information, where real-world sensor data updates the virtual model, and insights derived from the model inform and optimize the physical system’s operations \cite{anik5258720biotwinmine}.

DTs are increasingly applied across various domains, including healthcare \citep{nadeem2025comprehensive}, manufacturing \citep{fantozzi2025digital}, smart city \citep{judijanto2024trends} and aerospace \citep{garbarino2024digital}. Additionally, DTs are becoming essential in various other fields such as construction, where they enhance productivity, operational efficiency, and sustainability through the integration of data acquisition, processing, simulation, and decision support technologies \citep{mousavi2024digital}. For instance, DT systems in highway tunnel pavement performance prediction, using multiple time series stacking (MTSS), improve accuracy and timeliness in performance forecasting \citep{yu2020prediction}. Spatial DTs (SDTs) are also being built with key spatial technologies, categorized into four layers, which help enhance the accuracy and functionality of digital models \citep{ali2024enabling}. Furthermore, in the railway sector, systems like DefectTwin integrate multimodal and multimodel AI pipelines, improving defect detection accuracy and maintenance efficiency through real-time feedback and synthetic dataset generation \citep{ferdousi2024defecttwin}.

The ability to analyze and simulate complex interrelationships makes DTs particularly relevant in the field of pavement health monitoring, where real-time assessment of pavement conditions is critical for effective maintenance planning. 

\subsubsection{Relevance of Digital Twin in Pavement Health Monitoring}

In pavement health monitoring and maintenance optimization, DTs serve as an intelligent, centralized data hub that continuously updates based on sensor inputs, environmental conditions, and traffic data \citep{sun2024approach}. Real-time data from strain gauges, displacement sensors, temperature sensors, and moisture sensors are continuously used by DT technology to enable real-time performance tracking \citep{han2025proactive, hussein2024integrating}. Because of this constant information flow, pavement characteristics may be closely monitored, enabling engineers to spot early indicators of distress such as surface deformations, rutting, and cracks \citep{rasheed2020digital}. Preemptive testing of maintenance plans is made possible by the combination of simulation and predictive analytics, which guarantees that roadwork is scheduled with the least amount of disturbance to traffic flow \citep{bhatt2025architecting}. By choosing the least invasive and most efficient repair techniques, this strategy not only lowers maintenance costs but also increases pavement lifespan \citep{werbinska2024digital}. Table \ref{tab:dt_monitoring_aspects} shows an overview of the relevance of DT technologies in pavement health monitoring, highlighting key features, and potential benefits.

\begin{table}[ht]
\centering
\caption{Digital Twin in Pavement Health Monitoring}
\label{tab:dt_monitoring_aspects}
\resizebox{\textwidth}{!}{%
\begin{tabular}{|p{4cm}|p{8cm}|p{8cm}|}
\hline
\textbf{Aspect} & \textbf{What DT Monitors} & \textbf{Improvement by DT} \\ \hline

\textbf{Structural Health} & Cracks, rutting, and fatigue damage using real-time data and simulation models \citep{sun2024data} & Predicts structural failure and optimizes maintenance scheduling\\ \hline

\textbf{Thermal Behavior} & Temperature variations and heat accumulation through thermal sensors & Identifies thermal stress zones to guide material selection and layout \citep{menges2024predictive} \\ \hline

\textbf{Moisture \& Drainage} & Water infiltration patterns and drainage efficiency using hydrological sensing \citep{zhu2024methods} & Detects poor drainage conditions and suggests improvement strategies \\ \hline

\textbf{Traffic Load Impact} & Load distribution and stress-strain behavior from vehicle interactions \citep{rumpa2023infrachain} & Analyzes real-time stress conditions to inform reinforcement planning \citep{rumpa2023infrachain} \\ \hline

\textbf{Material Aging} & Asphalt oxidation and binder degradation using chemical and physical aging models \citep{sierra2022development} & Predicts degradation rates to optimize material renewal schedules \\ \hline

\textbf{Intelligent Sensing} & Data from embedded advanced sensors and IoT devices for continuous feedback  & Enhances data accuracy and monitoring efficiency via sensor fusion \citep{wang2024digital} \\ \hline

\textbf{Lifecycle Management} & Pavement’s full lifespan from construction to decommissioning \citep{torzoni2024digital} & Improves infrastructure planning and reduces lifecycle costs through proactive decision-making \\ \hline

\end{tabular}}
\end{table}

\subsection{Graph Neural Networks}

GNNs are a class of deep learning models specifically designed to perform inference on data structured as graphs. Their fundamental goal is to learn low-dimensional vector representations (embeddings) for nodes, edges, or entire graphs, capturing both the features of graph elements and the underlying graph topology \citep{wu2023graph, 4700287}.

\paragraph{Message-Passing in Graphs.}
Most GNN architectures operate based on a message-passing mechanism, where nodes iteratively update their representations by aggregating information from their neighbors and combining it with their own current representation \citep{Gilmer2020}. This process typically involves two main steps at each layer $k$ for a node $v$:

\textbf{Message Aggregation:} Information (messages) from the neighboring nodes $\mathcal{N}(v)$ is aggregated. A general form is:
\begin{equation}
    m_{\mathcal{N}(v)}^{(k)} = \text{AGGREGATE}^{(k)}\left(\left\{ h_u^{(k-1)} \mid u \in \mathcal{N}(v) \right\} \right)
\end{equation}
where $h_u^{(k-1)}$ is the representation of neighbor $u$ from the previous layer, and $\text{AGGREGATE}^{(k)}$ is a permutation-invariant function (e.g., sum, mean, max).

\paragraph{Update:} The aggregated neighborhood vector $m_{\mathcal{N}(v)}^{(k)}$ is combined with the node $v$'s own representation from the previous layer $h_v^{(k-1)}$ and transformed to produce the new representation:
\begin{equation}
    h_v^{(k)} = \text{UPDATE}^{(k)}\left(h_v^{(k-1)}, m_{\mathcal{N}(v)}^{(k)}\right)
\end{equation}
The $\text{UPDATE}^{(k)}$ function often involves a neural network layer. This iterative process allows information to propagate across the graph, enabling nodes to capture information from increasingly larger neighborhoods \citep{simran2025message}.

\begin{figure}
    \centering
    \includegraphics[width=\linewidth]{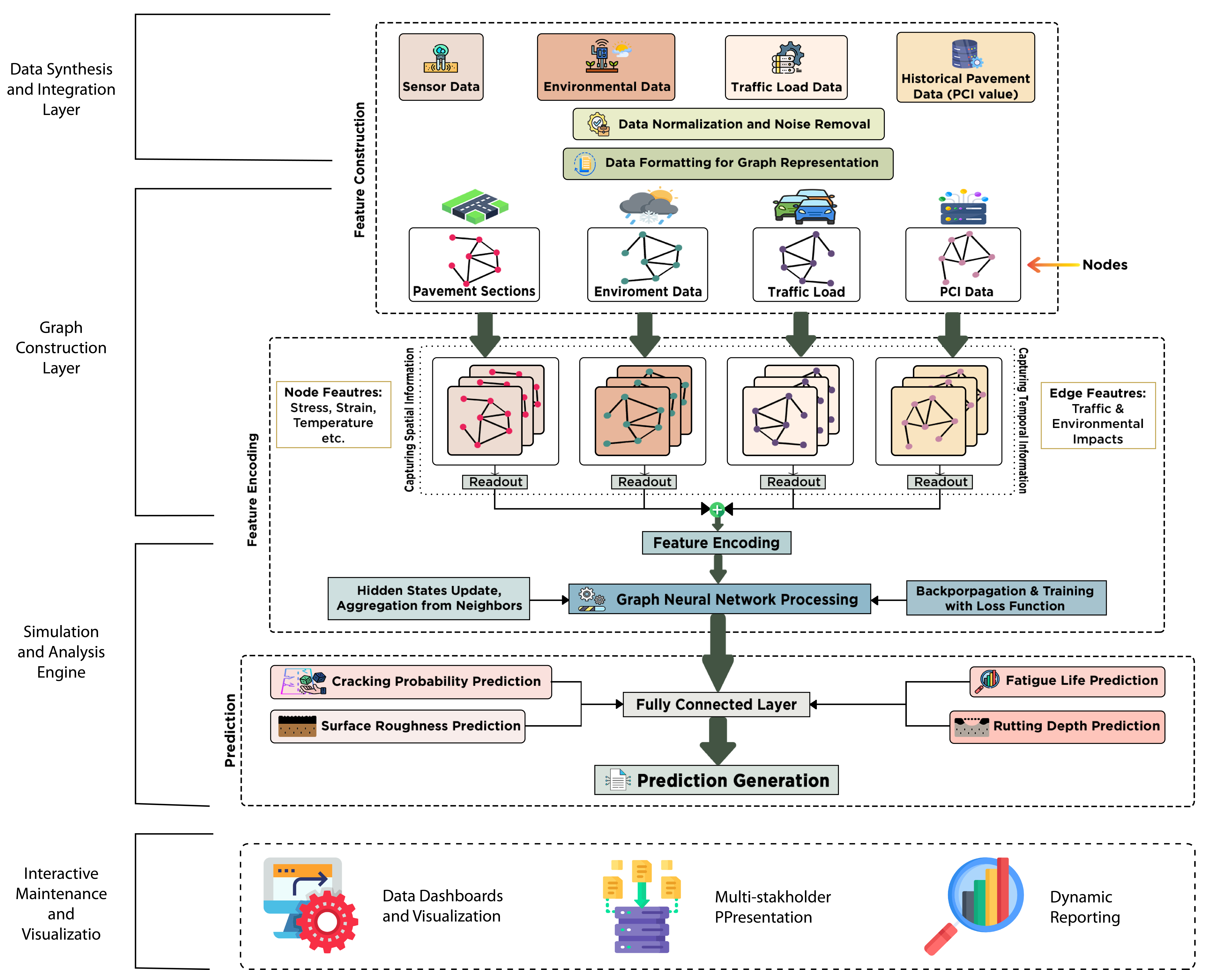}
    \caption{The Architecture of Graph Neural Network for Pavement Deterioration Prediction }
    \label{fig:Main2}
\end{figure}

Figure \ref{fig:Main2} illustrates this architecture, demonstrating how graph-structured input data flows through GNN layers to generate predictive insights related to pavement deterioration. The nodes in our graph representation correspond to individual pavement sections or segments within the monitored road network \citep{s22239183}. Each node signifies a distinct spatial unit of the pavement infrastructure, typically defined by geometric boundaries or maintenance zones. Each pavement section node is defined by a comprehensive feature vector that encapsulates various physical and structural parameters. The node features include stress and strain measurements, temperature readings, structural properties, and current condition indicators such as Pavement Condition Index (PCI) values \citep{WANG2024105480}.
The edges in the graph represent the spatial relationships and connectivity between adjacent pavement sections \citep{PETRASOVA2020104801}. These connections encode the physical adjacency of pavement segments and capture how deterioration patterns, environmental conditions, and traffic loads propagate through the pavement network. The edge features encapsulate the dynamic factors that influence the interaction between connected pavement sections. These features primarily include traffic load data, environmental impact parameters, and connectivity strength measures \citep{HE2023128913}. Traffic volume, load distribution, environmental conditions such as temperature fluctuations and precipitation, and the degree of influence between adjacent sections are encoded as edge attributes, allowing the model to capture how external factors affect pavement performance across spatial boundaries \citep{s22239183}.

In our pavement health monitoring system, each pavement section node generates messages by combining its own condition data with edge-specific factors, such as traffic and environmental conditions, and sends this information to its neighbors \citep{Gilmer2020}. Incoming messages from adjacent sections are aggregated, allowing each node to capture the health status of its local neighborhood \citep{wu2023graph}. Nodes then update their features by integrating these aggregated messages with their own state and historical data, capturing both spatial and temporal degradation patterns \citep{simran2025message}. Finally, the updated node representations are used to predict key pavement performance metrics, supporting proactive maintenance planning \citep{4700287}.

\section{Related Works} \label{sec:RelatedWorks}

In recent years, transportation infrastructure has seen significant advancements with the growing use of DT technologies and GNN in enhancing pavement monitoring and deterioration modeling. These innovations are transforming how pavements are managed, maintained, and optimized. In this section, we review the existing literature on the application of DTs in pavement health monitoring and maintenance, the use of GNN in pavement deterioration modeling, and the integration of GNN methods within DT frameworks. By examining the current research, we aim to highlight the gaps that still exist in this field and suggest potential directions for future work. Our goal is to further advance these approaches to effectively address the growing complexities of pavement monitoring and deterioration modeling in modern transportation infrastructure.

\subsection{Digital Twin Approaches in Pavement Monitoring}

The concept of DT is revolutionising the transportation industry by enabling precise infrastructure planning, operation, and maintenance. By creating highly accurate virtual models of transportation networks—including roads, railways, airports, and ports—DTs facilitate the effective and sustainable management of transportation assets \citep{yan2023digital}. This technology has been widely adopted across various fields such as aerospace \citep{jiang2022building}, industrial manufacturing \citep{dubarry2023enabling, liu2023systematic, zhang2023intelligent}, medical services \citep{aluvalu2023novel, cao2023digital}, agricultural machinery \citep{gallego2023sustainability, slob2023virtual}, electrical engineering \citep{biard2022reliability, sifat2023towards},  and urban management \citep{huang2022development,al2023pluralism}. Its application has increasingly expanded into transport infrastructure, offering innovative solutions for optimizing maintenance and operational strategies.

In the field of transport infrastructure, DTs have been employed to enhance the design and monitoring of various structures. For instance, \cite{konkov2023formation} developed a Finite Element Analysis-based DT model to evaluate load distribution on interstation tunnels in deep subway networks, streamlining metro tunnel design and reducing project timelines. \cite{ye2023digital} validated a practical DT application in the Dongtianshan Tunnel project in China, which successfully addressed extreme weather conditions, harmful gases, and complex geological challenges. Similarly, \cite{wang2022construction} conducted an extensive review of 145 studies on DT applications in underground infrastructure, spanning asset location mapping, construction coordination, and maintenance optimization. DTs are also gaining traction in bridge and highway monitoring. \cite{torzoni2022structural} leveraged DT technology to replicate bridge damage patterns, enabling structural health monitoring without the need for direct on-site measurements. DTs for operation and maintenance in the road sector have been introduced for example dealing with the maintenance of tunnels \citep{yu2021digital}, bridges \citep{lu2019digital,kaewunruen2021digital}, or road pavement \citep{bosurgi2020bim}. \cite{jiang2022building} introduced a novel method for creating highway DTs using specialized map data, optimizing asset management. \cite{kanigolla2024architecting} further demonstrated the effectiveness of DTs in real-time water distribution network optimization, showcasing their potential in predictive maintenance through data-driven simulations. In pavement engineering, \cite{bosurgi2020bim} highlighted DT functionalities such as real-time pavement condition surveys, interactive visualization of distress types, and geometric, structural, and functional quality assessments. \cite{oreto2022bim} extended these applications by integrating Life Cycle Assessment (LCA) into DT frameworks, enabling bidirectional data exchange between Building Information Modeling (BIM) platforms and LCA tools. \cite{yu2020prediction} further advanced pavement DTs by incorporating machine learning into BIM-based DT models for predictive performance analysis.

Despite significant advancements, scaling the adoption of DTs in pavement monitoring and maintenance remains challenging. Key barriers, such as data interoperability issues, real-time integration complexities, and the absence of standardized frameworks, continue to hinder widespread implementation \citep{barykin2021place}. The current state of Pavement Management Systems (PMS) also faces challenges related to fragmented data integration and limited predictive capabilities, underscoring the need for a comprehensive, technology-driven PMS framework \citep{wang2022bim}. Overcoming these limitations through integrated approaches and technological innovation is crucial to unlocking the full potential of DTs in optimizing pavement lifecycle management. 

\subsection{GNN-Centric Approaches in Pavement Deterioration Predictions and Maintenance}

Graph-based methodologies have gained prominence in pavement deterioration modeling due to their ability to capture the spatial dependencies inherent in road networks. Among these, GNNs have demonstrated superior performance in tasks such as pavement condition prediction, deterioration modeling, and maintenance scheduling, often surpassing traditional machine learning approaches. In our review, we found that studies have highlighted the effectiveness of GNNs in modeling complex relationships between pavement sections, leading to improved predictive accuracy and optimized maintenance strategies \citep{gao2024considering,lu2024graph}. The development of Graph Convolutional Networks (GCNs) by \cite{kipf2016semi} laid the foundation for GNN applications in pavement deterioration modeling. Subsequent advancements, such as Graph Attention Networks (GAT) \citep{velickovic2017graph} and GraphSAGE \citep{hamilton2017inductive}, further improved the scalability and robustness of these models.

In the context of pavement management, \cite{gao2024considering} introduced a convolutional GNN model for imputing missing pavement condition data, significantly outperforming conventional machine learning methods. Recent studies have explored the integration of GNNs with pavement monitoring frameworks. Traditional methods, such as edge detection \citep{li2018automatic} and threshold-based segmentation \citep{kamaliardakani2016sealed}, have given way to deep learning-based approaches. \cite{pan2018detection} demonstrated the effectiveness of Support Vector Machines (SVMs), Artificial Neural Networks (ANNs), and Random Forest models in detecting pavement cracks from multispectral UAV imagery. More recent GNN-based models have further enhanced predictive capabilities by integrating spatial dependencies within road networks. GNNs have also been utilized in time series forecasting for predicting pavement deterioration. \cite{singh2024spatial} introduced a novel fault diagnosis approach that combines GAT and long-short-term memory (LSTM) networks, improving the detection of nonlinear dependencies in pavement condition data. Similarly, \cite{gao2024considering} demonstrated that the incorporation of spatial dependencies in GNN-based pavement deterioration models significantly improves the predictive accuracy. Advancements in GNN-based structural health monitoring extend beyond pavement deterioration. \cite{zhou2024graph} developed a GNN simulator to model 3D pavement responses under tire loading, utilizing finite element (FE) simulations to capture dynamic behaviors. The study revealed that GNN-based simulations achieved high accuracy while significantly reducing computational costs compared to traditional FE models. These applications demonstrate how graph-based methodologies can substantially enhance the predictive accuracy of pavement deterioration models, optimize maintenance strategies, and improve the overall efficiency of pavement management systems, enabling more proactive and data-driven decision-making.

\subsection{Research Gap Analysis and Our Contribution}

Despite growing interest in DT and graph-based approaches in pavement management, their integration remains largely unexplored. Existing research focuses on isolated DT applications rather than holistic lifecycle implementations across design, construction, operation, and maintenance \citep{redelinghuys2020six, kulkarni2019towards}. Moreover, graph-centric approaches, especially GNNs, show promise in infrastructure management, such as monitoring pavement health, prediction of thermal fatigue, and rutting \citep{boonsiripant2024comparative}. However, their integration with DT frameworks remains underexplored. While some studies apply DTs in infrastructure monitoring and maintenance \citep{sun2024approach, jeon2024prescriptive}, graph-theoretical methods to model pavement interdependencies are lacking. Though DTs have been used to simulate pavement behavior under varying conditions \citep{boccardo2024urban}, few incorporate graph models to predict environmental impacts like temperature and traffic load. Unlike previous studies focusing on DT-driven asset management or independent GNN applications, our approach unifies them to enhance predictive accuracy, optimize maintenance strategies, and facilitate continuous monitoring, advancing pavement health management.

\section{Motivation} \label{sec:motivation}
Advancements in intelligent infrastructure management are reshaping pavement monitoring and deterioration modeling. DT and GNN have emerged as transformative tools that offer data-driven solutions to optimize maintenance and rehabilitation strategies \citep{chikwendu2024digital,aykurt2024digital}. DTs provide dynamic, real-time representations of infrastructure \citep{han2025proactive}, while GNNs improve predictive accuracy by using spatial and temporal relationships within pavement networks \citep{boonsiripant2024comparative}. However, pavement management still relies on fragmented methodologies, limiting the full potential of these technologies \citep{grilli2023methodologic}. This research explores the integration of DT and GNN to bridge existing gaps, enabling more efficient, predictive, and intelligent pavement lifecycle management.

\subsection{Limitations of Traditional Pavement Management Systems}

Traditional pavement management systems (PMS) have long relied on empirical models, periodic inspections, and manual assessments—a paradigm that often leads to delayed interventions \citep{tamagusko2024machine}, inconsistent data integration \citep{amandio2021integration}, and limited predictive capabilities \citep{talaghat2024digital}. In contrast, the emergence of DTs presents a promising evolution in the field. DTs enable real-time simulation and continuous monitoring of pavement conditions by integrating sensor-based data with historical records, thereby offering a dynamic tool for structural health monitoring \citep{sakr2024recent}. Despite these advances, the application of DTs in pavement management has been constrained by interoperability challenges, computational limitations, and the absence of standardized frameworks \citep{lu2025development}.

Parallel to these developments, predictive modeling stands as a significant challenge within PMS \citep{tamagusko2024machine}. Recent advancements in GNNs have attracted considerable attention due to their proficiency in modeling complex, interconnected systems—such as road networks—by treating individual pavement sections as graph nodes and capturing the inherent spatial dependencies \citep{ranu2024road}. Empirical studies have demonstrated that GNNs excel over traditional machine learning models in forecasting pavement distress, optimizing maintenance planning, and enhancing rehabilitation strategies \citep{radwan2025comparative}. However, these applications of GNNs are often fragmented, relying on static datasets that overlook real-time influences such as environmental conditions, traffic loads, and material aging \citep{longa2023graph}.

The synthesis of DTs and GNNs represents an unexplored frontier in pavement management. While research has independently validated the merits of DTs for infrastructure monitoring \citep{jayasinghe2024application} and GNNs for predictive modeling \citep{majidiparast2025graph}, there remains a notable absence of an integrated framework that harnesses the strengths of both approaches. Such a framework could transform pavement lifecycle management by enabling real-time data processing, predictive maintenance, and automated decision-making. Bridging this gap can revolutionize pavement lifecycle management by enabling real-time data processing, predictive maintenance, and automated decision-making.

\subsection{Combined Benefits of DT and GNN}

The integration of DTs and GNNs represents a paradigm shift in pavement management, offering unparalleled advantages over traditional methodologies. The ability to visualize structural performance in a virtual model allows engineers to test various intervention strategies before implementing them in real-world scenarios, thereby improving cost-effectiveness and decision-making \citep{azanaw2024revolutionizing,lemian2025digital}. Complementing the predictive capabilities of DTs, GNNs offer a powerful approach to modeling complex pavement deterioration processes \citep{ranu2024road}. 

For this research, the GNN will be employed due to its ability to adaptively capture spatial dependencies (pavement connectivity, traffic loads, and material properties) by assigning different importance levels to neighboring pavement segments \citep{vrahatis2024graph}. GNNs dynamically learns which pavement regions contribute most to deterioration prediction, ensuring more precise assessments \citep{wei2024neighbor}. By integrating DT and GNN, a real-time feedback loop will be established. The DT continuously updates pavement conditions, which the GNN model then analyzes to predict future deterioration. These insights are used to optimize maintenance schedules, determining when and where interventions should be prioritized to prevent severe damage, minimize repair costs, and extend pavement lifespan. The combination of real-time data analysis, predictive modeling, and automated decision making ensures an intelligent, data-driven approach to monitoring and optimizing pavement maintenance \citep{cai2025engineering,razavi2023deep, wasi2025theoreticalframeworkgraphbaseddigital}. The convergence of these technologies holds the potential to transform traditional PMS into a proactive, data-driven system, ensuring safer, more resilient, and cost-effective infrastructure networks for the future. Table \ref{tab:dt_gnn_pavement_advancements} shows key advancements in pavement infrastructure monitoring and maintenance achieved through the integration of Digital Twin and GNN technologies

\begin{table}[ht]
\centering
\caption{Advancements in Pavement Infrastructure Monitoring and Maintenance with DT \& GNN}
\label{tab:dt_gnn_pavement_advancements}
\resizebox{\textwidth}{!}{%
\begin{tabular}{|p{5cm}|p{6.5cm}|p{8.5cm}|}
\hline
\textbf{Aspect} & \textbf{Current Approach} & \textbf{Potential Benefits of DT \& GNN} \\ \hline

\textbf{Pavement Condition Assessment} & Manual inspections, sensor-based monitoring, and image analysis techniques \citep{ifeanyi2024graph} & DT enables real-time condition updates, while GNN enhances predictive accuracy for surface and subsurface deterioration \\ \hline

\textbf{Predictive Maintenance} & Traditional rule-based strategies and statistical forecasting models & GNN-driven predictive modeling reduces premature interventions and optimizes maintenance cycles \citep{habibollahi2023predicting} \\ \hline

\textbf{Traffic and Environmental Impact Analysis} & Empirical models based on historical traffic and climate data \citep{oakley2023foresight} & DT captures real-time traffic and weather impacts; GNN improves the forecasting accuracy of pavement wear and damage \\ \hline

\textbf{Cost Optimization} & Conventional cost-benefit analysis using static historical maintenance cost data & DT-GNN integration enables dynamic cost modeling and supports long-term investment strategies \citep{vallarino2024dynamic} \\ \hline

\end{tabular}}
\end{table}

\section{Framework Overview} \label{sec: framework-emthodology}

\begin{figure}[htbp]
    \centering
    \includegraphics[width=\linewidth]{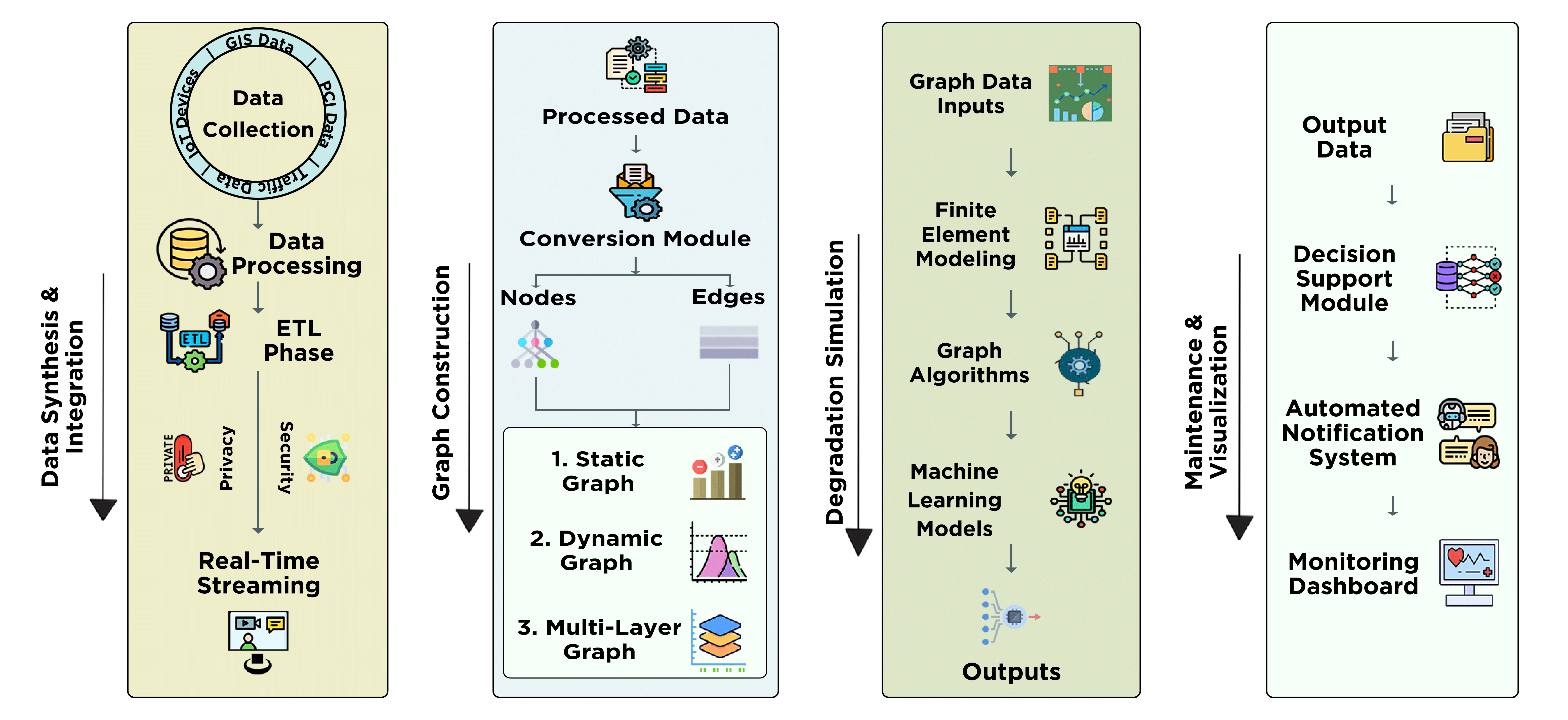}
    \caption{Detailed framework of graph-based Digital Twins for pavement health monitoring and maintenance.}
    \label{fig:Main3}
\end{figure}

In this section, we detail the architecture of a Graph-based DT (GDT) specifically developed for monitoring pavement health and guiding maintenance strategies. Figure \ref{fig:Main3} illustrates the full architecture, which integrates real-time data from embedded sensors, traffic monitoring systems, GIS data, and maintenance records to form a comprehensive digital representation of pavement infrastructure. The architecture is organized into multiple layers—from data ingestion and preprocessing to graph construction and dynamic analysis—each playing a pivotal role in ensuring accurate, timely, and actionable insights for pavement performance and maintenance planning.

\subsection{Data Synthesis and Integration Layer}

\begin{figure}[htbp]
    \centering
    \includegraphics[width= \linewidth]{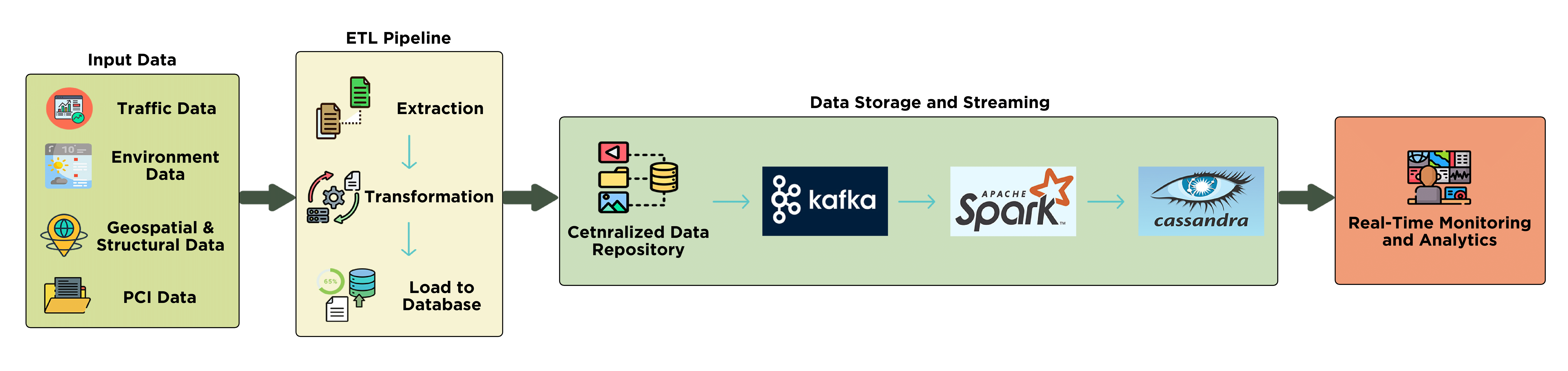}
    \caption{Schematic Representation of Data Integration Layer Processes and Components}
    \label{fig:Main4}
\end{figure}

The Data Integration Layer is critical to our DT architecture, ensuring the seamless aggregation, standardization, and preprocessing of heterogeneous data sources into a unified model. This layer collects real-time measurements from a variety of sources, ranging from embedded pavement sensors to external environmental data—enabling continuous monitoring and proactive maintenance decision-making \citep{werbinska2024digital}. Figure \ref{fig:Main4} illustrates a schematic representation of the key processes and components involved in this data integration layer. By implementing a robust data integration strategy, this layer forms the foundational basis upon which the entire DT’s decision-support capabilities are built \citep{amandio2021integration}. The ability to accurately harmonize diverse data streams ensures that subsequent analytical modules function with high reliability and minimal propagation of uncertainty.

\subsubsection{Data Sources} 
To comprehensively capture the diverse dimensions of pavement health, several critical data contributors are integrated into this layer. Traffic data remains essential, with datasets sourced from highway monitoring systems and Weigh-in-Motion (WIM) sensors. These data streams offer insights into axle loads, vehicle classifications, and traffic volume patterns, all of which significantly impact pavement distress mechanisms. Such metrics are vital for computing Equivalent Single Axle Loads (ESALs) and evaluating dynamic impact factors that drive structural degradation, including fatigue and rutting \citep{zhao2022mechanistic}. Environmental data is another crucial component. Sourced from meteorological stations, satellite-based remote sensing, and IoT-enabled weather sensors, these datasets offer a comprehensive view of environmental influences on pavement behavior. Factors such as temperature fluctuations, variations in relative humidity, precipitation intensity, and solar radiation directly interact with pavement materials, affecting their chemical and structural integrity \citep{barisic2021thermal}. Long-term environmental data allows for the development of climate-specific degradation models, refining the DT’s predictive accuracy based on geographic location \citep{yan2024deep}.

Equally important is the collection of pavement condition data. These datasets reflect the physical health of the surface and include high-frequency structural assessments, UAV-based LiDAR imaging, infrared thermography, and in-situ sensor networks \citep{bennet2008data}. Such sources yield real-time indicators of surface distress, including cracking, rutting, stripping, and roughness. These measurements not only offer a snapshot of the pavement’s current state but also enable predictive modeling by establishing time-series deterioration curves \citep{gao2023deep}. Complementing the above datasets are geospatial and structural data. Derived from GIS databases, pavement management systems, and engineering design records, these datasets cover material compositions, layer thicknesses, subgrade properties, and historical maintenance interventions. Incorporating this structural information enables the DT to simulate how pavement design responds to external stressors. Furthermore, it facilitates the construction of spatial dependency matrices, which are essential for defining graph topologies within the analytical components of the system \citep{gao2024considering}.

\subsubsection{Data Preprocessing}
Following data acquisition, all inputs undergo rigorous preprocessing to ensure they are accurate, consistent, and complete \citep{liu2023systematic}. This process is anchored in a tailored Extract, Transform, Load (ETL) pipeline designed for complex pavement-related datasets \citep{searls2020systems}.

The extraction phase initiates the process by gathering raw data from embedded sensors, traffic systems, PMMS databases, and public weather APIs. The ETL system is equipped to manage various data formats and communication protocols to ensure that no critical information is lost. During this stage, sensor calibration metadata is also recorded to correct for device-specific biases and allow comparability between devices \citep{kumaran2021etl}.

Next, in the transformation phase, the raw data is cleaned, normalized, and synchronized. Outlier detection techniques such as Z-score normalization and Mahalanobis distance filtering are applied to eliminate anomalies and improve data consistency \citep{rajamani2025enhancing}. The transformation process addresses issues such as missing values, duplicate records, and inconsistencies in measurement units. For example, sensor data recorded in varying units is standardized, while temporal mismatches are resolved to generate coherent time-series datasets across pavement segments. A multi-resolution time alignment mechanism further harmonizes high-frequency sensor readings with lower-frequency inspection reports, promoting analytical uniformity \citep{maharana2022review}.

Finally, the load phase concludes the preprocessing pipeline. The transformed data are stored in a centralized high-throughput database optimized for real-time retrieval and analytics \citep{searls2020systems}. This data repository supports downstream graph construction and complex machine learning applications \citep{biswas2022modeling}. The system is configured for efficient data loading to enable near-real-time updates. Moreover, metadata tagging is applied at every transformation step to ensure traceability, facilitate audits, and maintain backward compatibility with future versions of the DT system \citep{michals2022building}.

\subsubsection{Real-Time Streaming and Processing Frameworks}

To handle the demands of real-time data ingestion and processing in pavement health monitoring, we utilize advanced data streaming frameworks. Technologies like Apache Kafka facilitate scalable, fault-tolerant, and low-latency data handling, making them ideal for continuously integrating high-frequency sensor data from UAV-based LiDAR scans, weather sensors, and WIM-enabled pavement monitoring systems \citep{vyas2021literature}. Kafka efficiently processes large streams of environmental and structural data, ensuring seamless ingestion and reducing the risk of data loss or delays. Its distributed architecture allows horizontal scaling across regional monitoring nodes, ensuring resilience and uninterrupted data flow under peak loads \citep{garg2013apache}. To enhance Kafka’s capabilities, we integrate it with Apache Spark for real-time data transformation and Cassandra for efficient NoSQL storage, creating a robust data pipeline for predictive analysis \citep{salloum2016big, abramova2013nosql}. This architecture supports complex event processing (CEP), enabling near-instantaneous detection of anomalies or threshold breaches in pavement condition metrics. This setup allows real-time cleaning, enrichment, and modeling of incoming data, such as detecting pavement distress, crack propagation, or structural weaknesses \citep{chaudhari2019scsi}.

\subsubsection{Challenges}
Implementing the Data Integration Layer for pavement health monitoring presents several challenges that must be addressed for effective system performance. Data heterogeneity poses a significant hurdle, as integrating continuous sensor streams, discrete maintenance records, and environmental data requires robust standardization protocols to ensure consistency \citep{guerrero2014scalable}. Incomplete or missing data can disrupt analysis, often caused by sensor outages or communication failures \citep{li2019handling}. Interoperability issues arise due to varying data formats and communication protocols across different systems, necessitating middleware solutions and standardized data schemas for seamless integration \citep{petrasch2022data}. Furthermore, temporal misalignment between asynchronous data streams can distort spatiotemporal inference, requiring advanced imputation algorithms and temporal resampling strategies \citep{luu2021time}. Security and privacy require secure data handling and regulatory compliance, necessitating scalable cloud-based infrastructures to manage complexity without compromising performance or increasing costs \citep{Ezeugwa_2024}.

\subsection{Graph Construction}

The Graph Construction Module transforms the integrated and preprocessed data from the Data Integration Layer into a dynamic graph representation that models the pavement network. This module serves as the foundation for advanced analysis and decision-making by capturing both the structural and temporal interdependencies across the pavement infrastructure.

\subsubsection{Graph Nodes and Edges}
In our graph model, the nodes represent critical pavement segments or maintenance zones, each enriched with attributes that define pavement health. Mathematically, the set of nodes is defined as \( {N} = {n_1, n_2, ..., n_k} \), where each node \({n_i}\) encapsulates key operational metrics necessary to assess pavement performance.
Edges in the graph model represent the interactions and dependencies between pavement segments, capturing key relationships that influence deterioration patterns. The set of edges is defined as \( E = {e_1, e_2, ..., e_m} \), with each edge \({e_i}\) annotated with attributes such as load transfer coefficients, connectivity strength, and temporal delay factors.
Table \ref{tab:graph_types_usecases} provides a classification of different graph types, outlining their purposes, key features, and typical applications relevant to pavement infrastructure.

\begin{table}[ht]
\centering
\caption{Classification of Graph Types with Purposes, Features, Applications, and Illustrative Use Cases}
\label{tab:graph_types_usecases}
\resizebox{\textwidth}{!}{%
\begin{tabular}{|p{4cm}|p{5cm}|p{5cm}|p{6cm}|}
\hline
\textbf{Graph Type} & \textbf{Primary Purpose} & \textbf{Characteristics} & \textbf{Representative Use Cases} \\ \hline

\textbf{Static Graphs} & Provide a fixed structural overview of the network at a specific point in time \citep{wang2025survey} & Emphasize network topology, structural bottlenecks, and node connectivity \citep{singh2024gtagcngeneralizedtopologyadaptive}& Identify deteriorated pavement segments disrupting optimal traffic flow \citep{zhou2020graph} \\ \hline
\textbf{Dynamic Graphs} & Model temporal changes in road conditions, traffic behavior, or system responses \citep{10.1145/3308560.3316581} & Incorporate time-varying attributes such as traffic load or climate effects \citep{10.1145/3308560.3316581} & Simulate the impact of seasonal traffic surges on pavement deterioration \citep{khemani2024review} \\ \hline
\textbf{Multi-Layer Graphs} & Represent heterogeneous relationships across interconnected infrastructure and operational layers \citep{liu2024multidimensional} & Distinct layers for structural data, environmental factors, and usage patterns & Evaluate how delayed maintenance (financial layer) affects surface distress propagation \citep{khemani2024review} \\ \hline

\end{tabular}}
\end{table}

\subsubsection{Graph-Based Representations of Pavement Networks}
Each graph type serves a distinct analytical purpose. Static graphs facilitate baseline diagnostics and benchmarking, while dynamic graphs enable temporal forecasting and anomaly detection. Multi-layer graphs support systemic insights across interdependent domains of pavement behavior \citep{tong2025stgan}. The selection and integration of these representations are vital to capturing the pavement's holistic lifecycle behavior under multifactorial influences \citep{wettewa2024graph}.

\paragraph{\textbf{Static Graphs.}}
Static graphs provide a snapshot of the pavement network at a specific moment, capturing its fixed structure and interdependencies. They are instrumental in identifying bottlenecks and pinpointing segments with poor condition or limited load-bearing capacity that may require immediate attention \citep{wang2025survey}. This static representation serves as a foundation for assessing long-term pavement conditions and planning targeted interventions.
A static graph is mathematically represented as  \(G = (V, E)\), where \(V\) denotes the set of pavement segments and \(E\) represents the fixed connections between them \citep{zhou2020graph}.

\paragraph{\textbf{Dynamic Graphs.}}
Dynamic graphs enhance static representations by integrating temporal variations, capturing how pavement conditions evolve over time due to weather, traffic loads, and wear. This approach enables real-time monitoring, allowing continuous tracking of deterioration processes and assessing the effectiveness of maintenance actions \citep{10.1145/3308560.3316581}.
Dynamic graphs are modeled as \(G(t) = (V(t), E(t))\), where the attributes of nodes and edges are functions of time, capturing the evolving state of the pavement network \citep{khemani2024review}.

\paragraph{\textbf{Multi-Layer Graphs.}}

Multi-layer graphs offer a comprehensive framework for modeling pavement networks by organizing critical aspects of pavement health into interconnected layers \citep{liu2024multidimensional}. The structural layer represents the physical integrity and connectivity of pavement segments, ensuring that load-bearing capacity and material properties are accurately captured. The surface distress layer focuses on visible defects such as cracks, potholes, and rutting, providing insights into surface-level deterioration. The environmental layer integrates external influences, such as temperature, precipitation, and solar radiation, which significantly impact pavement aging and degradation. Meanwhile, the traffic load layer maps dynamic load patterns and stress distributions, highlighting how vehicular movement affects pavement performance over time.
Each layer is modeled as \(G_i = (V, E_i)\), and the multi-layer graph is constructed by interconnecting these layers through shared nodes \citep{khemani2024review}.

\begin{figure}[htbp]
    \centering
    \includegraphics[width=0.9 \linewidth]{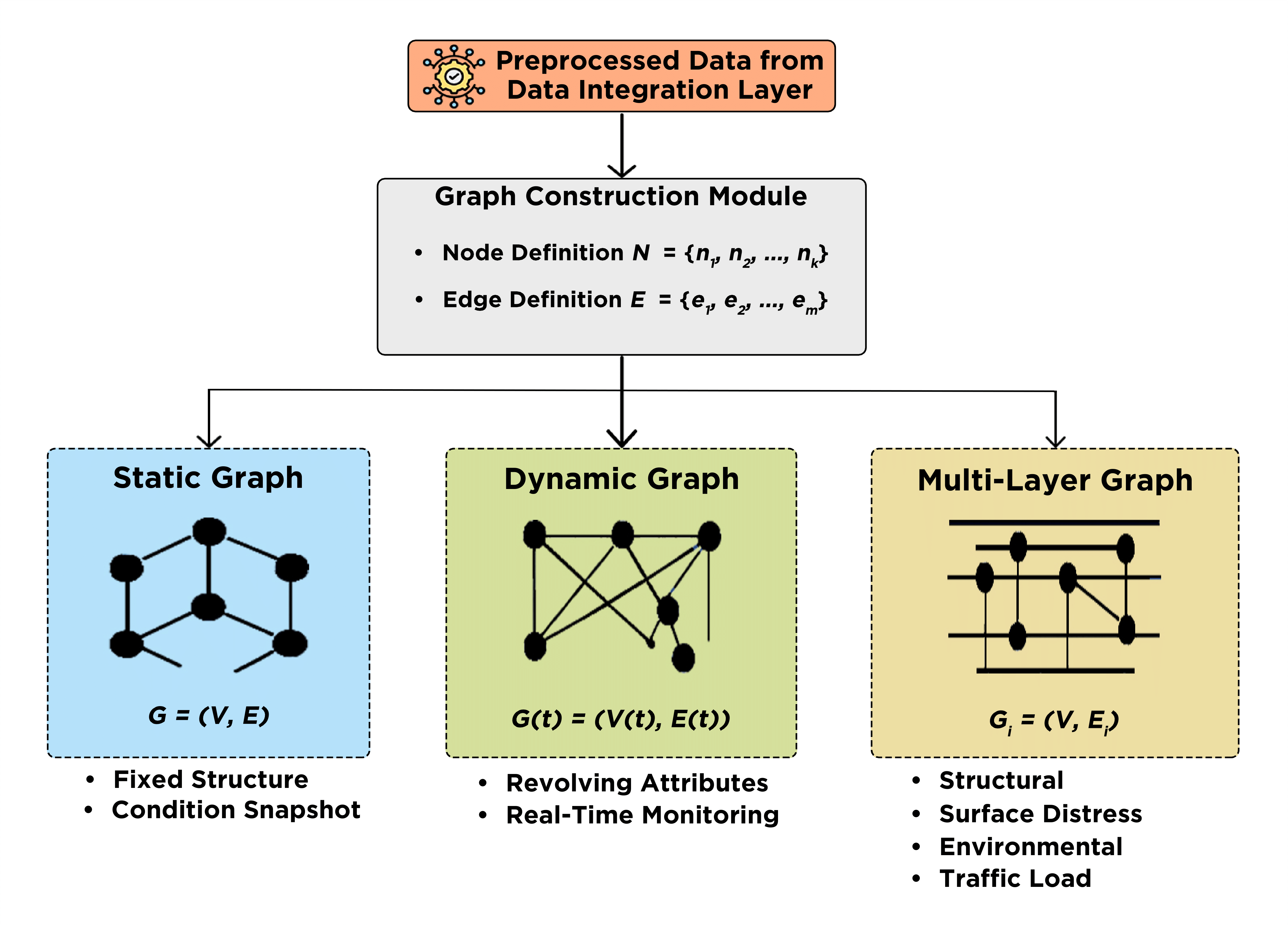}
    \caption{Workflow and Components of Graph Construction Layer}
    \label{fig:Main5}
\end{figure}

The principal objective of the Graph Construction Module is to systematically convert the harmonized and preprocessed data derived from the Data Integration Layer into a sophisticated, weighted graph that encapsulates the intricate spatial-temporal and functional interdependencies governing pavement systems. This module is strategically designed to transcend the limitations of conventional, linear, and compartmentalized asset management models by offering a holistic, multidimensional representation of pavement network components and their dynamic interactions. Figure \ref{fig:Main5} illustrates the workflow and key elements of this graph construction layer, highlighting how complex relationships are encoded into a graph-based structure for subsequent analysis and interpretation.

\subsection{Simulation and Analysis Engine}
\begin{figure}[htbp]
    \centering
    \includegraphics[width= \linewidth]{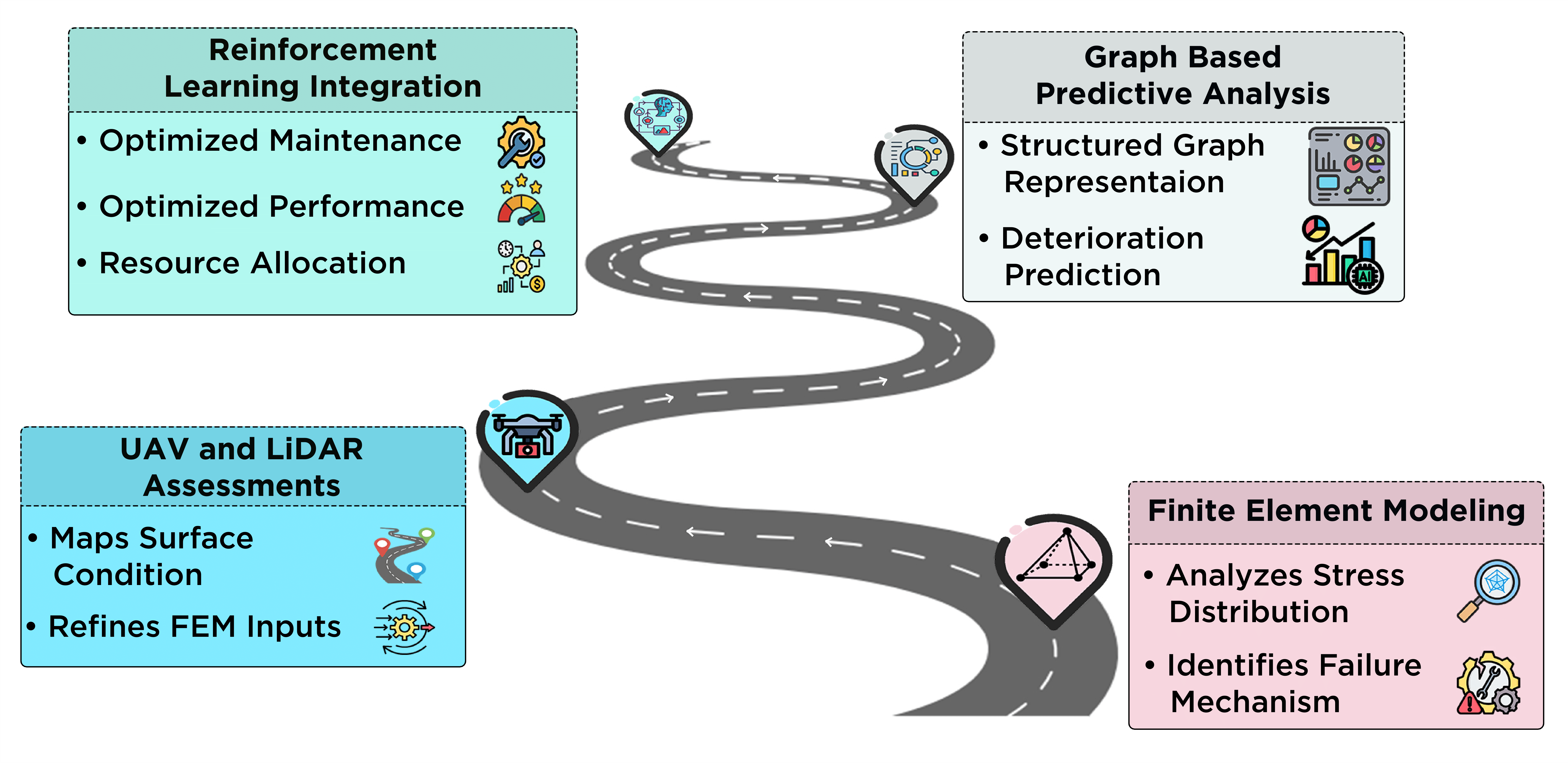}
    \caption{Framework for Integrated Pavement Simulation and Optimization}
    \label{fig:Main6}
\end{figure}

At the core of the proposed framework lies the Simulation and Analysis Engine. It serves as an integrated module that synthesizes multiple advanced computational methodologies to capture pavement degradation phenomena and accurately inform optimized maintenance strategies. Figure \ref{fig:Main6} presents a schematic overview of the workflow and components involved in the degradation simulation framework. It is meticulously designed to simulate the dynamics of pavement degradation, evaluate the effectiveness of various maintenance interventions, and facilitate the optimization of repair schedules in a proactive and data-driven manner.

\paragraph{\textbf{Finite Element Modeling:}}
The framework initiates with Finite Element Modeling (FEM), a foundational technique that breaks down pavement structures into discrete elements to analyze internal stress distribution, thermal effects, and fatigue behavior under varying vehicular and environmental conditions \citep{charhi2023modeling}. This method enables a granular evaluation of failure mechanisms, such as rutting and fatigue cracking, thereby informing early interventions to prevent systemic degradation. The granular insights from FEM serve as critical input parameters for subsequent predictive and decision-making modules \citep{assogba2020finite}.

\paragraph{\textbf{UAV and LiDAR-Based Surface Assessments:}}
To augment the fidelity of FEM simulations, high-resolution surface condition mapping is conducted using Unmanned Aerial Vehicles (UAVs) and LiDAR scanning technologies \citep{jankauskiene2020surface}. These non-intrusive assessments provide detailed topographic and structural insights, refining FEM inputs through accurate detection of surface distresses such as cracks, rutting, and delamination \citep{hackney20152}. The integration of UAV and LiDAR ensures spatially continuous and temporally updated pavement diagnostics, which are essential for dynamic model calibration \citep{li2024high}.

\paragraph{\textbf{Graph-Based Predictive Analysis:}}
Building upon structural and surface data, the framework transitions to a Graph-Based Predictive Analysis module. Within this graph, nodes represent infrastructural elements- pavement segments, sensors, intersections—enriched with high-dimensional features, while edges define their functional and physical relationships \citep{wettewa2024graph}. Predictive algorithms, including GNNs, are applied to model degradation trajectories and enable forward-looking maintenance planning \citep{wang2025survey}. The use of community detection and centrality measures allows for the identification of critical zones and interconnected vulnerabilities across the pavement network \citep{nippani2023graph}.

\paragraph{\textbf{Reinforcement Learning Integration:}}
The final stage integrates Reinforcement Learning (RL) algorithms, including Deep Q-Networks (DQNs) and Multi-Agent Reinforcement Learning (MARL), to dynamically optimize maintenance decision-making processes \citep{silva2019reinforcement}. These algorithms learn adaptive strategies for repair scheduling, resource distribution, and maintenance routing based on evolving pavement conditions and real-time traffic patterns \citep{santos2019adaptive}. RL integration enables proactive and goal-oriented interventions that minimize life-cycle costs and enhance serviceability \citep{asghari2022reinforcement}.

\subsection{Interactive Maintenance and Visualization}
\begin{figure}[htbp]
    \centering
    \includegraphics[scale=0.21]{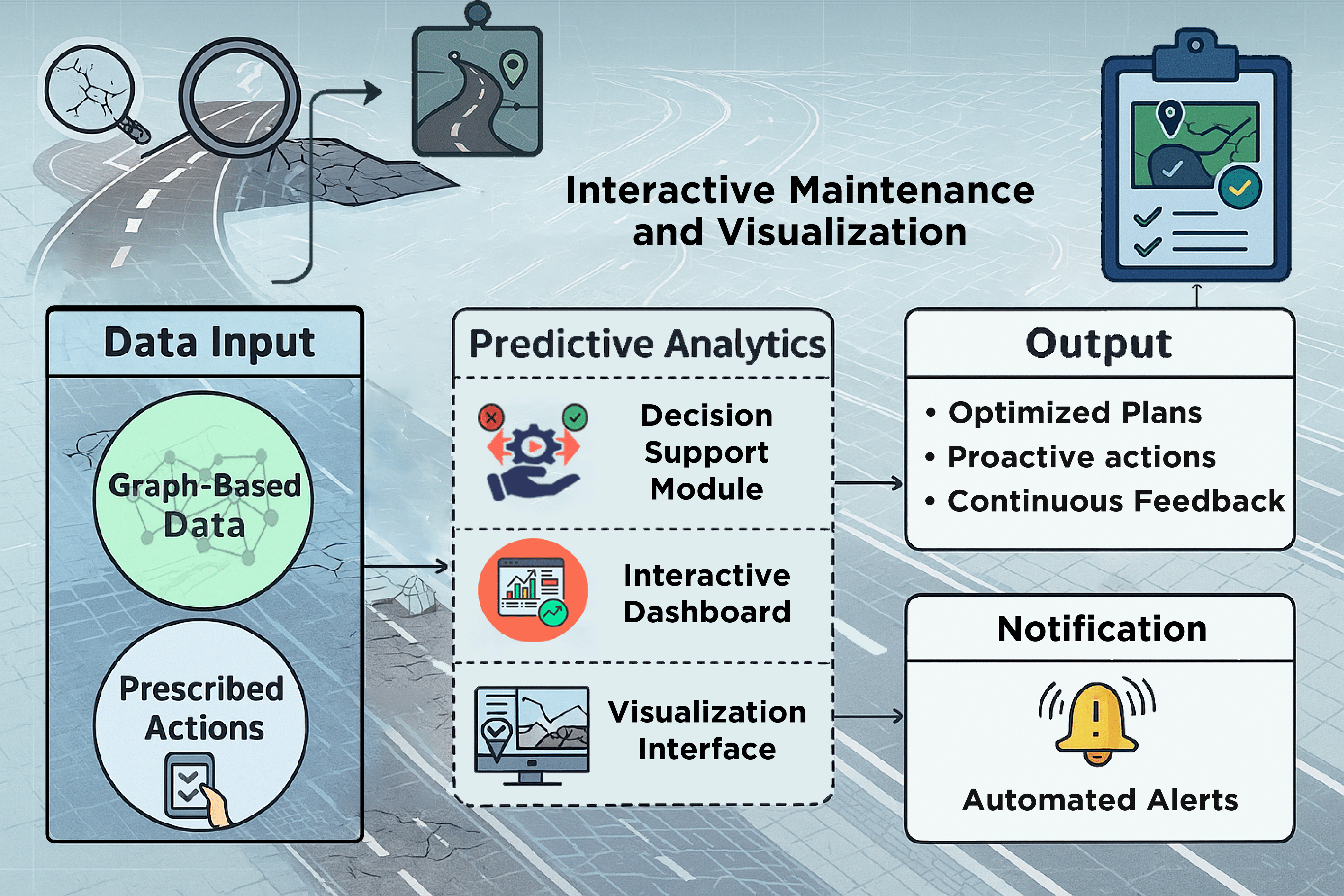}
    \caption{Interactive Maintenance and Visualization System}
    \label{fig:Main7}
\end{figure}

The Interactive Maintenance and Visualization System (IMVS) serves as a next-generation graph-driven DT framework designed to support proactive pavement maintenance informed by data \citep{del2018augmented}. As illustrated in Figure \ref{fig:Main7}, the system architecture is centered around three interconnected components: data input, predictive analytics, and intelligent outputs—culminating in optimized maintenance planning and real-time responsiveness. The process begins with data input, where real-world conditions are captured through diverse sources and encoded into graph-based representations of the pavement network. These data sources encompass historical maintenance records, prescribed rehabilitation actions, traffic loads, sensor measurements, and environmental stressors. Such integration of heterogeneous information enables the system to represent structural and functional interdependencies within the pavement ecosystem \citep{makendran2024designing}. This input is then processed through a robust predictive analytics engine, which encompasses a Decision Support Module, an Interactive Dashboard, and a Visualization Interface \citep{wajiddigital}. The Decision Support Module acts as the analytical core of the system, interfacing with traditional Pavement Management Systems (PMS) while extending their capabilities through machine learning and statistical inference models \citep{tamagusko2024machine}. It generates key performance indicators—such as the Pavement Condition Index (PCI), International Roughness Index (IRI), and thermal degradation profiles—based on temporal and spatial data encoded within the graph structure. The Interactive Dashboard translates analytical outputs into real-time, user-friendly visual formats, allowing stakeholders to explore pavement conditions at multiple layers of granularity \citep{kalamaras2017interactive}. This dashboard supports cross-comparative evaluations, highlights distressed segments, and visualizes predicted deterioration trajectories over time. Adjacent to this, the Visualization Interface provides advanced geospatial rendering and scenario simulations, offering stakeholders a comprehensive understanding of the long-term consequences of varied maintenance strategies \citep{petrasova2020geospatial}.

\begin{table}[ht]
\centering
\caption{Key Feedback Components in Pavement Digital Twin Systems}
\label{tab:feedback_loop_pavement}
\resizebox{\textwidth}{!}{%
\begin{tabular}{|p{4cm}|p{5cm}|p{5cm}|p{6cm}|}
\hline
\textbf{Component} & \textbf{Function} & \textbf{DT Role} & \textbf{Example} \\ \hline

\textbf{Model Conformance} & Validates simulations against design standards \citep{charhi2023modeling}. & Verifies structural and graph models \citep{assogba2020finite}. & Aligning FEM outputs with lab-tested pavement responses. \\ \hline

\textbf{Predictive Assessment} & Compares forecasted and actual degradation \citep{jankauskiene2020surface}. & Refines deterioration modeling \citep{hackney20152}. & Updating crack models using UAV/LiDAR data \citep{li2024high}. \\ \hline

\textbf{Real-Time Integration} & Ingests live sensor and traffic data. & Maintains model fidelity over time \citep{wettewa2024graph}. & Recalibrating models with environmental sensor inputs. \\ \hline

\textbf{Discrepancy Analysis} & Detects simulation-performance gaps. & Adjusts model parameters and edge weights \citep{wang2025survey}. & Tuning GNNs based on observed vs. predicted distress \citep{nippani2023graph}. \\ \hline

\textbf{Algorithm Tuning} & Applies learning-based refinement \citep{silva2019reinforcement}. & Enhances scheduling and interventions \citep{santos2019adaptive}. & Using RL to optimize future maintenance plans \citep{asghari2022reinforcement}. \\ \hline

\end{tabular}}
\end{table}

The system’s actionable insights are operationalized through an Automated Notification Module, which continuously monitors infrastructure performance and triggers automated alerts \citep{karthick2024ai} when threshold breaches are detected, such as critical PCI drops or abrupt crack propagation events. These alerts facilitate timely interventions and resource reallocation before significant deterioration occurs. The final output layer of the system generates a suite of tangible deliverables, including optimized maintenance plans, proactive mitigation actions, and continuous feedback loops. Table \ref{tab:feedback_loop_pavement} highlights the key feedback components integral to Pavement Digital Twin systems, detailing their functions, and roles in enabling real-time monitoring and adaptive decision-making.

\section{Experiments} \label{sec: experiments}
In this section, we present the experimental setup used to evaluate the performance of our graph-based approach for pavement condition prediction. We outline the dataset characteristics, preprocessing steps, model configurations, and implementation details.
\subsection{System Architecture}

We develop a graph-based learning system for pavement distress estimation that captures structural dependencies and physical attributes of road networks through inductive representation learning. The architecture models segments as nodes and connectivity as edges, enabling effective feature aggregation over the graph topology derived from real-world pavement data.

\begin{figure}[htbp]
    \centering
    \includegraphics[scale=0.70]{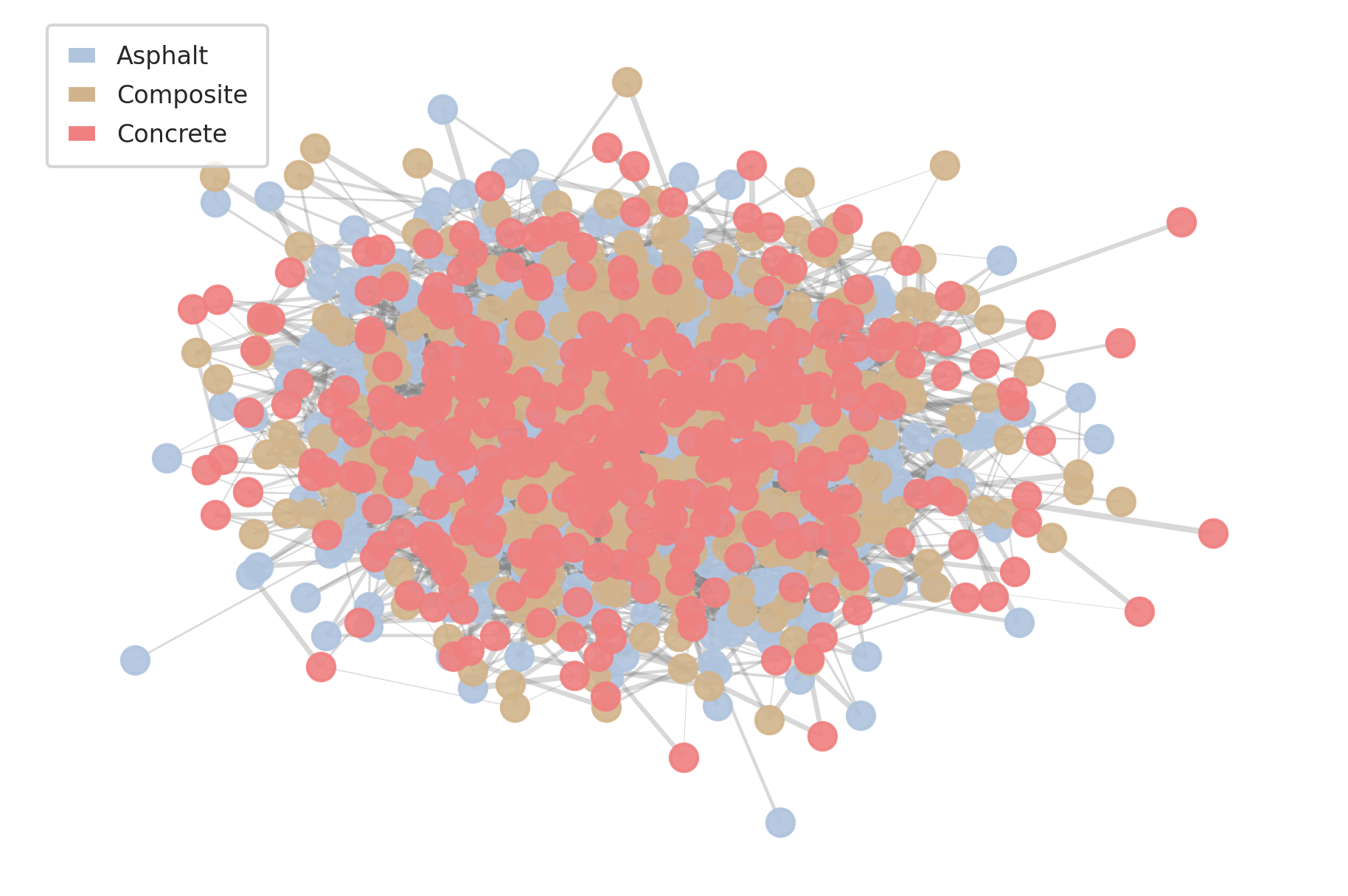}
    \caption{Graph representation of pavement segments (nodes) and their connectivity (edges) based on segment IDs.}
    \label{fig:pave_graph}
\end{figure}

The input comprises three data sources: (i) segment-level pavement attributes (e.g., length, material, age, traffic), (ii) temporally evolving distress records, and (iii) a directed connectivity graph denoting adjacent segments. We use the most recent distress observation per segment as the supervision target, while node features are standardized and embedded into a continuous feature space. The resulting network structure, where pavement segments are represented as nodes and physical connections as edges, is visualized in Figure~\ref{fig:pave_graph}. This heterogeneous graph captures spatial dependencies across road segments and serves as the foundation for graph-based message passing in our model.

Let $\mathcal{G} = (\mathcal{V}, \mathcal{E})$ be the undirected pavement graph with $|\mathcal{V}| = N$ segments and $|\mathcal{E}| = E$ bidirectional edges. Each node $i$ has a feature vector $\mathbf{x}_i \in \mathbb{R}^s$ representing local characteristics, and the initial node feature matrix is $\mathbf{X} \in \mathbb{R}^{N \times s}$. During learning, each node updates its representation via a two-stage neighborhood aggregation function. The representation of node $i$ at layer $\ell$ is given by:

\begin{equation}
\mathbf{h}_i^{(\ell)} = \mathbf{W}_1 \mathbf{h}_i^{(\ell-1)} + \mathbf{W}_2 \cdot \text{AGG}_{j \in \mathcal{N}(i)} \left( \mathbf{h}_j^{(\ell-1)} \right)
\end{equation}

where $\mathcal{N}(i)$ denotes the set of neighboring nodes of $i$, $\mathbf{W}_1, \mathbf{W}_2 \in \mathbb{R}^{d \times d}$ are learnable weights, and AGG is an aggregation operator (e.g., mean pooling). This formulation facilitates both ego-node transformation and neighborhood information aggregation, supporting inductive learning on previously unseen graph segments.

To improve model expressiveness, a non-linear activation is applied between layers, and the final node output is computed after two such message-passing iterations. The model architecture includes two aggregation layers with ReLU activations and a single scalar output predicting segment-level distress:

\begin{equation}
\hat{y}_i = f(\mathbf{h}_i^{(2)}) \in \mathbb{R}
\end{equation}

To ensure generalization and avoid information leakage, a random 80/20 split of nodes is used for training and evaluation, and dropout regularization is applied during training. The model is optimized using Adam with a small learning rate and $L_2$ weight decay.

This inductive learning framework enables efficient and scalable prediction of pavement deterioration while respecting the topological and physical constraints of infrastructure networks. The system is robust to dynamic updates in graph structure, making it well-suited for real-time infrastructure monitoring tasks.

\subsection{Data Description}

We utilize a modified version of the DVRPC Pavement Condition dataset \citep{dvrpc_pavement_conditions}, adapted to support graph-based modeling. The original tabular data was structurally and semantically augmented into three components: Pavement Segment Data, Distress Data, and Connectivity Data. This enriched representation enables spatiotemporal learning and is made available as part of our supplementary materials.

The Pavement Segment Data defines each road segment with attributes such as \texttt{segment\_id}, length, material, age, and traffic volume. The Distress Data simulates monthly degradation levels (analogous to PCI), capturing dynamic evolution influenced by traffic, material aging, and environmental wear. The Connectivity Data encodes pairwise relationships among segments—representing physical or functional adjacency—and assigns weights to capture link strength. A heterogeneous graph was constructed using NetworkX, where nodes represent segments with associated features and the most recent distress level serves as the regression target. Edges were formed from the connectivity dataset, yielding a static graph that preserves both structural topology and current pavement condition.

\subsubsection{Data Preprocessing}

A systematic preprocessing pipeline was employed to ensure data consistency, integrity, and model readiness. Missing values were handled using median imputation for numerical fields and mode imputation for categorical features. Pavement material was encoded numerically using label encoding, and all numeric inputs were standardized via z-score normalization to ensure feature parity.

\begin{figure}[htbp]
    \centering
    \includegraphics[scale=0.6]{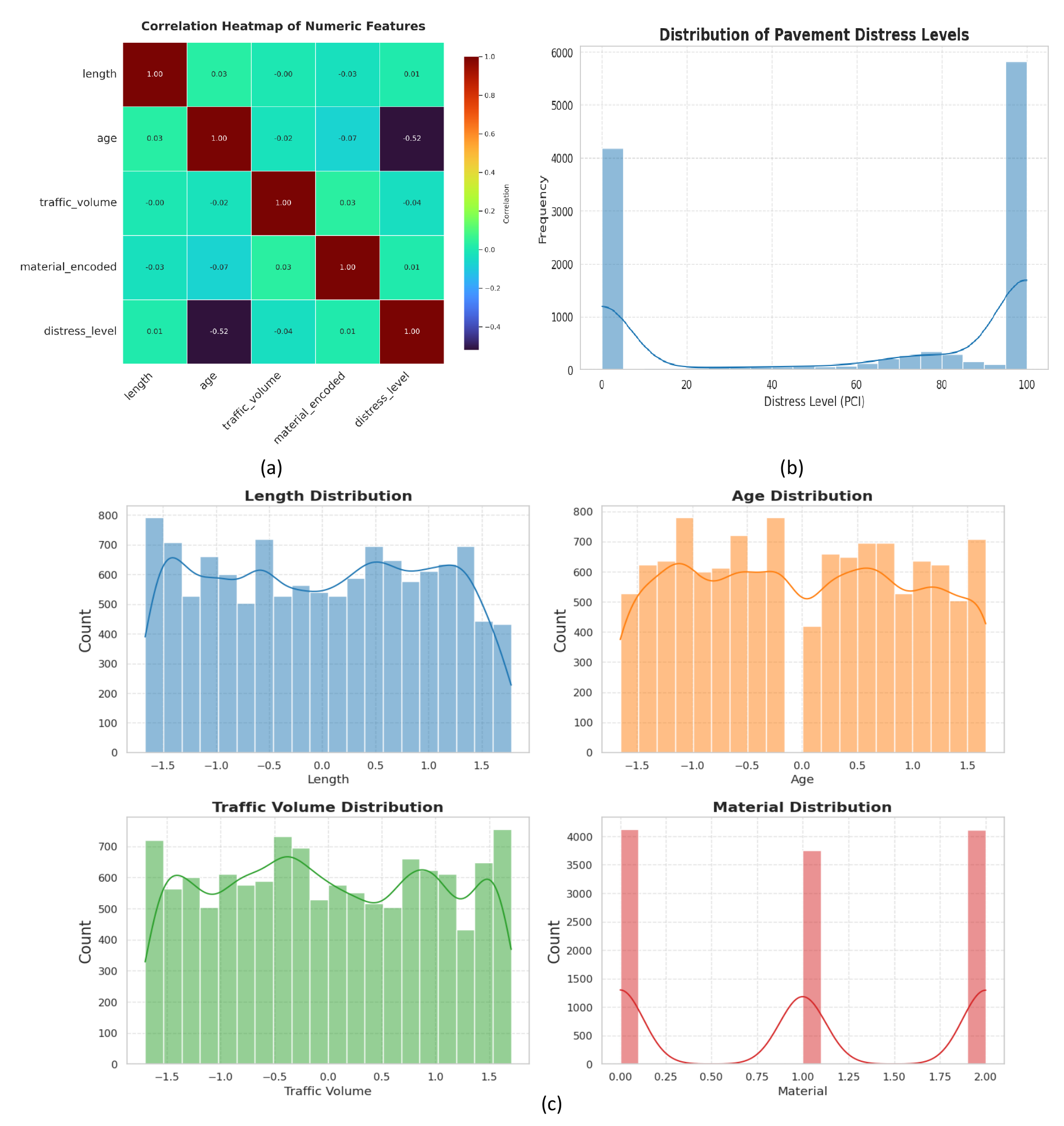}
    \caption{Exploratory data analysis visualizations. (a) Feature correlation structure; (b) Distribution of pavement distress levels (PCI) highlighting the overall condition variability across segments; (c) Distributions of key pavement segment features: length, age, traffic volume, and material type, illustrating their variability and contribution to node characterization in the graph model.}
    \label{fig:eda_combined}
\end{figure}

Exploratory data analysis (Figure~\ref{fig:eda_combined}) was conducted to assess feature distributions, skewness, and inter-variable correlations, guiding the final feature selection. The connectivity dataset was cleaned to remove invalid or incomplete links. The final graph structure was converted into PyTorch Geometric format: node attributes were transformed into float tensors, and the edge list was expressed as an \texttt{edge\_index} tensor in COO format. To support bidirectional message passing, reverse edges were explicitly added. The target variable, \texttt{distress\_level}, was reshaped as a floating-point tensor for node-level regression. This end-to-end pipeline ensured compatibility with GNN training while preserving domain-specific spatial dependencies.

\subsection{Baselines}

To benchmark the effectiveness of our graph-based model, we evaluated a suite of traditional machine learning regressors that operate exclusively on tabular node-level features, without leveraging graph structure. These models provide a reference point for assessing the added predictive value of incorporating topological context via GNNs. The baseline models include: Random Forest Regressor \citep{segal2004machine}, Gradient Boosting Regressor \citep{friedman2001greedy}, Linear Regression \citep{seber2003linear}, Support Vector Regressor (SVR) \citep{smola2004tutorial}, K-Nearest Neighbors Regressor (KNN) \citep{kramer2013k}, and Decision Tree Regressor \citep{xu2005decision}, all implemented using Scikit-learn \citep{pedregosa2011scikit}. Each model was trained on the same set of standardized node features—segment length, pavement age, traffic volume, and encoded material type—ensuring uniform input representation across baselines. Feature scaling was performed using z-score normalization. An 80/20 train-test split was applied with a fixed \texttt{random\_state=42} to ensure reproducibility \citep{jain2022diagnostic}. Model performance was evaluated using three standard regression metrics: Mean Absolute Error (MAE), Root Mean Squared Error (RMSE), and the coefficient of determination ($R^2$) \citep{plevris2022investigation}. 

These baselines serve to quantify the extent to which graph-aware learning improves predictive accuracy in scenarios where spatial connectivity and inter-node dependencies are relevant to the target variable.

\subsection{Implementation Details}

The model was trained using an 80/20 train-test node split with fixed random seeds (42) applied across NumPy \citep{oliphant2006guide} and PyTorch \citep{ketkar2021introduction} for reproducibility. The optimizer used was Adam, with a learning rate of 0.001 and L2 regularization via weight decay of $1 \times 10^{-5}$. Dropout (rate = 0.2) was applied after ReLU activation in the first hidden layer. The model was initially trained for 200 epochs and extended to 5000 epochs to ensure convergence. Mean Squared Error (MSE) was used as the loss function, aligning with the regression objective of distress level prediction. Hyperparameters were manually tuned through iterative experiments, varying learning rate, hidden dimensions, weight decay, and dropout, guided by validation loss and generalization performance. The model’s sensitivity to structural sparsity and synthetic noise underscored the importance of domain-specific tuning. The implementation leveraged PyTorch and PyTorch Geometric, with data processing handled via pandas, NumPy, and NetworkX. Baseline models (Random Forest, SVR, Gradient Boosting) were implemented using Scikit-learn. All experiments were conducted in a CPU-only environment (16 GB RAM), and runtime efficiency was achieved through sparse graph structures and mini-batch computation in PyTorch Geometric. Reproducibility was ensured through consistent preprocessing, fixed random seeds, and version-controlled code and datasets, which will be made available for benchmarking.

\subsection{Result and Analysis}
This section presents the outcomes of our experimental evaluation and provides a detailed analysis of the model’s performance. 

\subsubsection{Quantitative Performance}
To evaluate the predictive performance of the proposed GNN model, we employed three standard regression metrics: Mean Absolute Error (MAE), Root Mean Squared Error (RMSE), and the coefficient of determination ($R^2$). Each metric provides a different perspective on the model's accuracy and generalization ability.

\begin{itemize}
    \item \textbf{Mean Absolute Error (MAE)}: 
    \[
        \text{MAE} = \frac{1}{n} \sum_{i=1}^{n} \left| y_i - \hat{y}_i \right|
    \]
    This measures the average magnitude of absolute errors between predicted ($\hat{y}_i$) and true ($y_i$) values.

    \item \textbf{Root Mean Squared Error (RMSE)}:
    \[
        \text{RMSE} = \sqrt{ \frac{1}{n} \sum_{i=1}^{n} (y_i - \hat{y}_i)^2 }
    \]
    This penalizes larger errors more heavily, offering insight into prediction volatility.

    \item \textbf{Coefficient of Determination ($R^2$ Score)}:
    \[
        R^2 = 1 - \frac{ \sum_{i=1}^{n} (y_i - \hat{y}_i)^2 }{ \sum_{i=1}^{n} (y_i - \bar{y})^2 }
    \]
    This reflects the proportion of variance in the target variable that is captured by the model.
\end{itemize}

The GNN model was trained for 2000 epochs on a graph containing 1,000 nodes and 6,000 undirected edges, with four input features per node. The training loss steadily declined over epochs, indicating improved learning stability, with the final training loss reaching 963.08. However, the test loss remained comparatively higher at 1515.68, suggesting some degree of overfitting or limited generalization on the unseen data.

\begin{table}[htbp]
\centering
\caption{GNN Model Performance on Test Set}

\begin{tabular}{|c|c|c|c|}
\hline
\textbf{MAE} & \textbf{RMSE} & \textbf{$R^2$ Score} & \textbf{Test Loss (MSE)} \\
\hline
31.34 & 38.93 & 0.3798 & 1515.68 \\
\hline
\end{tabular}
\end{table}

These results indicate that the GNN effectively learned from both the node features and the graph structure, achieving moderate predictive accuracy under a synthetic pavement network setting.

\subsubsection{Comparative Evaluation}

To contextualize the performance of the proposed GNN model, we benchmarked it against several widely used regression models implemented using Scikit-learn. These include Random Forest, Gradient Boosting, Linear Regression, Support Vector Regression (SVR), K-Nearest Neighbors (KNN), and Decision Tree Regressor. Each model was trained on the same feature set—length, age, traffic volume, and encoded material type—and evaluated on identical train-test splits to ensure a fair comparison.

\begin{table}[htbp]
\centering
\caption{Performance Comparison of GNN vs. Traditional Models}
\begin{tabular}{|l|c|c|c|}
\hline
\textbf{Model} & \textbf{MAE} & \textbf{RMSE} & \textbf{$R^2$ Score} \\
\hline
GNN & 31.34 & 38.93 & 0.3798 \\
Random Forest & 36.2150 & 44.8784 & 0.1862 \\
Gradient Boosting & 37.2642 & 45.2680 & 0.1720 \\
Linear Regression & 36.3744 & 43.3392 & 0.2411 \\
SVR & 37.1333 & 47.4599 & 0.0899 \\
K-Nearest Neighbors & 37.8000 & 47.6655 & 0.0820 \\
Decision Tree & 31.5000 & 56.1249 & -0.2727 \\
\hline
\end{tabular}
\label{tab:model_comparison}
\end{table}

\noindent \textbf{GNN Performance:} The GNN model achieved the highest $R^2$ score (0.3798), indicating superior overall variance explanation compared to traditional models. Although its MAE (31.34) and RMSE (38.93) are not the lowest among all models, the results reflect a balanced trade-off between prediction accuracy and generalizability, especially under the presence of graph-based dependencies.

\noindent \textbf{Decision Tree Behavior:} The Decision Tree model exhibited the lowest MAE (31.5000), but its RMSE (56.1249) and negative $R^2$ score (-0.2727) suggest severe overfitting. This model fits training data closely but fails to generalize, performing poorly on unseen instances.

\noindent \textbf{Ensemble Models:} Random Forest and Gradient Boosting models demonstrated consistent and robust performance across all metrics. Their ensemble mechanisms allow them to capture complex feature interactions and reduce overfitting, making them strong baselines—even without graph-aware mechanisms.

\noindent \textbf{Linear and Kernel-Based Models:} Linear Regression offered a reasonable baseline with moderate performance, highlighting some linear relationships in the feature space. SVR and KNN regressors performed worse than ensembles and GNN, reflecting limited capacity to capture the full complexity of the data.

\begin{figure}[htbp]
    \centering
    \includegraphics[width=\textwidth]{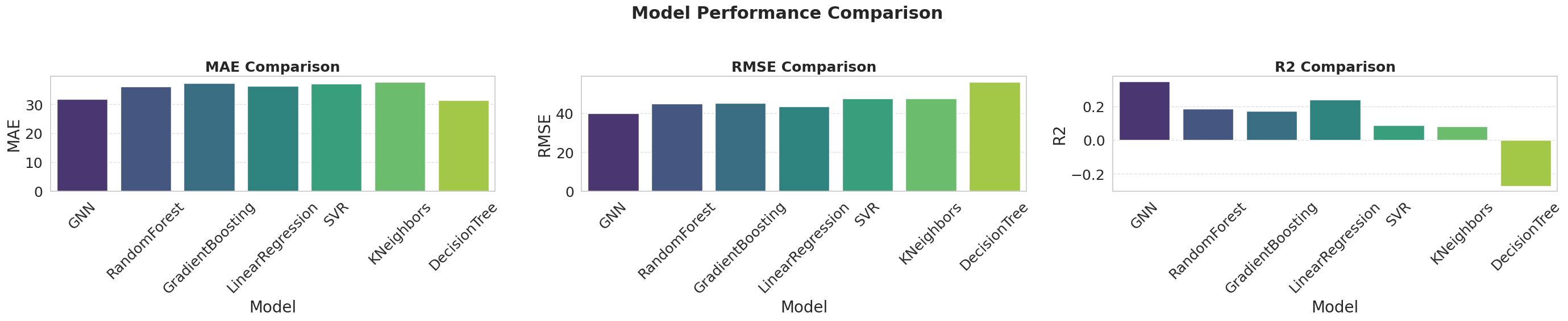}
    \caption{Comparison of MAE, RMSE, and R2 performance metrics across various machine learning models..}
    \label{fig:model_performance_compariso}
\end{figure}

\noindent Overall, while GNN outperforms in terms of $R^2$ score, the margin of improvement is modest. This suggests that in the current synthetic dataset, individual node-level features dominate predictive power, with limited incremental benefit from the structural information embedded in the graph.

\begin{figure}[htbp]
    \centering
    \includegraphics[width=\textwidth]{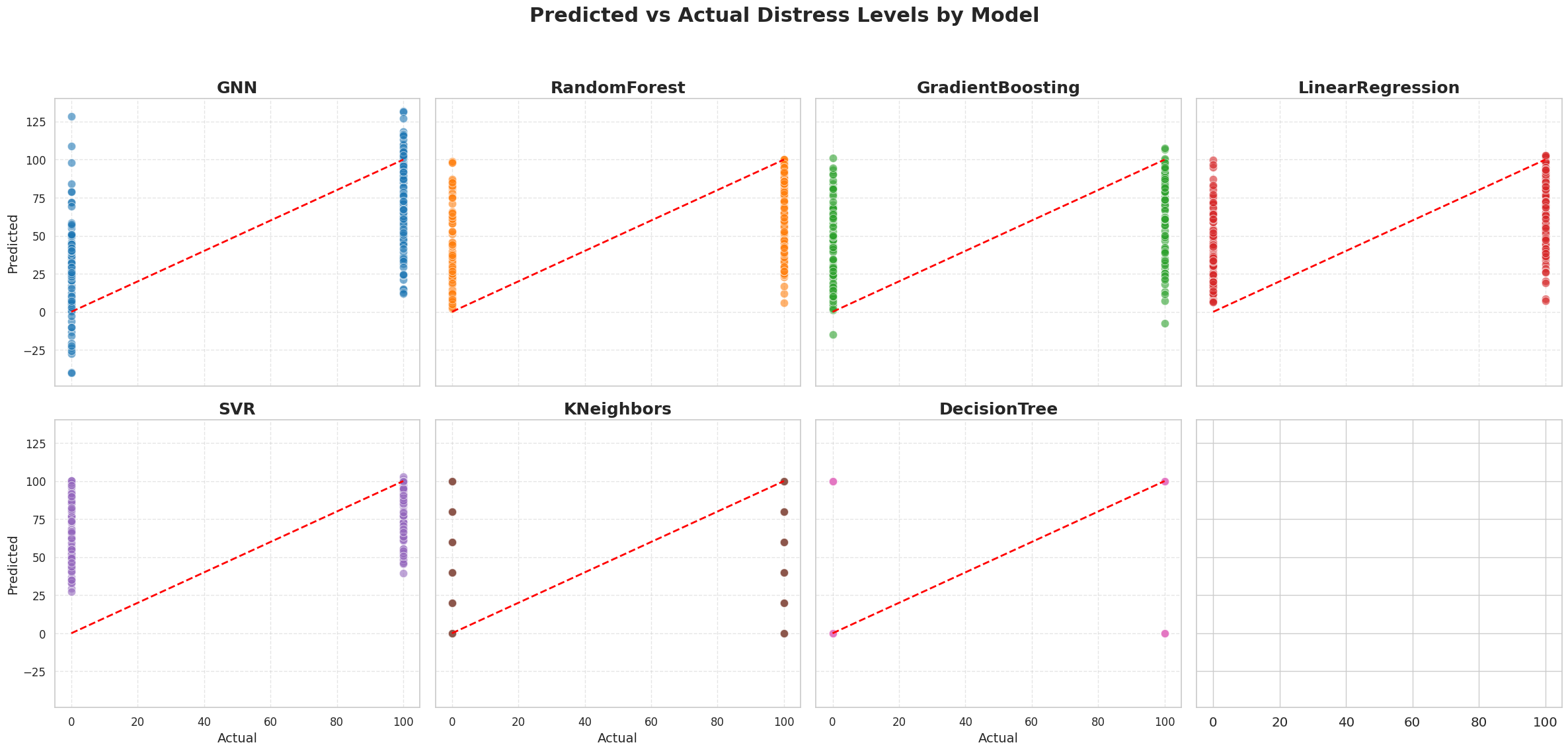}
    \caption{Scatter plots of predicted vs. actual distress levels for all model.}
    \label{fig:all_predictions_vs_actual}
\end{figure}

To further illustrate these differences, Figure~\ref{fig:all_predictions_vs_actual} visualizes the relationship between actual and predicted pavement distress levels across all models. Each subplot presents a scatter plot, where the x-axis represents actual distress (PCI), and the y-axis shows predicted values. The red dashed line ($y = x$) denotes the ideal prediction line. The scatter plot reveals distinct predictive behaviors among the models. The GNN model exhibits wider variance, particularly at the extremes of the PCI range, reflecting its ability to capture non-linear dependencies, though with reduced precision in certain regions. Random Forest and Gradient Boosting models tend to cluster more tightly in the upper PCI range but show diminished accuracy for lower values. Linear Regression and SVR yield smooth and continuous predictions yet underfit the higher and lower PCI values, indicating a limitation in capturing more complex patterns. In contrast, KNeighbors and Decision Tree models display discrete, step-like prediction behavior with noticeable clustering, indicative of localized overfitting and poor generalization. These visual insights underscore the nuanced trade-offs among the models, reaffirming the GNN’s capacity to leverage structural relationships in data, while highlighting its limitations under certain conditions.

\section{Discussion} \label{sec:discussion}

The integration of GNNs within the architecture of a DT represents a paradigm shift in the domain of pavement infrastructure management, offering a powerful approach to understanding, predicting, and optimizing pavement performance in real time \citep{Wang2024}. Traditional pavement management models have often struggled with the complexity and dynamism inherent in road networks, particularly in capturing spatial heterogeneity and temporal evolution \citep{huang2024developing}. However, the synergy between GNNs and DTs enables the modeling of pavements as intelligent, interconnected systems that closely mirror physical reality. By representing road networks as graphs where pavement segments, intersections, and embedded sensors function as nodes and edges, GNNs can effectively learn and utilize the spatial and temporal dependencies within the infrastructure \citep{wu2023graph}. Unlike conventional machine learning models that rely on flat or tabular data structures, GNNs can incorporate the topological structure of transportation systems, allowing the model to understand how deterioration in one segment may influence or correlate with adjacent segments \citep{ifeanyi2024graph}. This structural awareness becomes especially valuable in urban networks where traffic flow, environmental stressors, and maintenance activities are intricately interlinked \citep{tamagusko2024machine}.

The performance evaluation of the proposed GNN model further supports its relevance in practical implementation. Compared to a suite of traditional regression models—including Random Forest, Gradient Boosting, Linear Regression, SVR, KNN, and Decision Tree—the GNN achieved the highest $R^2$ score (0.3798), demonstrating stronger explanatory power for pavement distress variability. While some models like Decision Tree showed slightly lower MAE, their negative $R^2$ and high RMSE indicated overfitting and poor generalization. In contrast, the GNN model demonstrated a balanced trade-off between predictive accuracy and generalizability, particularly due to its ability to exploit graph-based dependencies. This robustness is further illustrated in the scatter plots comparing predicted vs. actual pavement condition values across models. The GNN model, though showing broader variance at the extremes, effectively captured non-linear patterns in distress evolution—an essential trait for modeling real-world infrastructure behavior. Moreover, the error analysis revealed that while the model is generally unbiased with a symmetrical error distribution centered around zero, it occasionally suffers from large deviations, likely stemming from edge cases or underrepresented classes. Notably, the confusion matrix indicated accurate classification in Low and High distress categories, though it struggled with moderate levels, possibly due to class imbalance or feature overlap. In the proposed theoretical framework, the DT acts as a continuously evolving digital replica of the physical pavement system, dynamically updated through live sensor inputs and historical maintenance records. Within this framework, GNNs serve as the core engine for predictive analytics, learning latent patterns from large-scale, high-dimensional datasets—such as crack propagation behavior, axle load distributions, thermal fluctuations, and moisture infiltration \citep{gao2024considering}. As these models evolve, they can accurately forecast pavement distresses, identify high-risk segments, and recommend optimal intervention timelines. This predictive capability forms the basis for condition-based maintenance, replacing inefficient reactive or periodic strategies with interventions that are precisely timed and targeted.

One of the major advantages of this integrated GNN-DT architecture lies in its ability to support closed-loop decision-making. Feedback from real-world performance is continuously compared against model forecasts, enabling recalibration of predictions and refinement of maintenance strategies \citep{singh2021digital}. This adaptive learning cycle ensures that the system improves over time, aligning maintenance planning with actual field conditions \citep{fuller2020digital}. The implementation of such feedback loops can significantly reduce lifecycle costs by minimizing over-maintenance and preventing catastrophic failures through early intervention \citep{narayanan2024machine}. Another important insight is the potential for this framework to enhance existing Pavement Management Systems (PMS). Traditional PMS often relies on heuristic rules or deterministic models, which are insufficient in the face of dynamic urbanization, climate change, and increasing traffic volumes \citep{maheshwari2024co}. By embedding GNN-based intelligence within the DT layer, PMS can evolve into cognitive systems that incorporate not only historical knowledge and engineering judgment but also real-time learning and adaptive response \citep{shahzad5085883cognitive}. This represents a convergence of infrastructure engineering and artificial intelligence, with the potential to inform both operational decisions and long-term strategic planning. Finally, from a broader perspective, the proposed approach aligns with emerging trends in smart infrastructure and digital transformation in civil engineering. As urban systems become increasingly complex, the ability to synthesize data-driven insights with domain expertise becomes crucial.

\subsection{ Applied Perspectives} \label{sec:Challenges-and-research-directions}
Graph-based Digital Twins (GDTs) are increasingly essential in pavement management, enabling real-time monitoring, predictive maintenance, and system-wide optimization \citep{Wang2024}. These models replicate road networks by integrating IoT sensors, traffic data, and environmental inputs to reflect actual conditions and forecast failures.

Ensuring pavement resilience is crucial amid rising traffic, aging infrastructure, and climate variability. Traditional monitoring often misses dynamic stressors like floods and overloads, leading to hidden deterioration \citep{braunfelds2022road}. GDTs address this by modeling interconnected networks that simulate disruptions and identify weaknesses early \citep{sierra2022development, ayvaz2021predictive}. A smart city in China used such a system with IoT sensors to detect weather-related pavement distress and act promptly during monsoon seasons \citep{yan2024digital}. Conventional maintenance relies on periodic checks, often causing inefficient resource use and uneven road conditions \citep{braunfelds2022road}. GDTs support continuous monitoring and decision-making by treating road networks as dynamic graphs, with machine learning predicting optimal repair times and interventions. A European authority using GDTs with GNNs reduced maintenance costs by 20 percent and improved the Pavement Condition Index across its highways \citep{Zhu2024DigitalTwinPavement, wang2024digital}.

GDTs also promote sustainability by analyzing material use, emissions, and energy consumption across the pavement lifecycle \citep{HUANG2009197, ma16031047}. A Dutch pilot study compared lifecycle emissions of flexible and rigid pavements using GDT simulations, guiding authorities to adopt greener designs in high-traffic corridors \citep{su151410908, Jin2021DevelopmentDigitalTwin}. Real-time condition monitoring is vital for proactive pavement management \citep{Wang2024}. Manual inspections often miss fast-developing issues, especially in high-stress zones \citep{infrastructures9070101}. GDTs, using embedded sensors and mobile data, allow early detection of distresses like cracking or settlement \citep{infrastructures5020018}. A U.S. state DOT used such a system to monitor thousands of miles of highways, enabling rapid anomaly detection and emergency rerouting during extreme weather events \cite{Shtayat2024MethodsDigitalTwinPavement}. Table \ref{tab:graph_dt_pavement_applications} summarizes current applications and highlights the broad potential of this integrated approach beyond the present study.

\begin{table}[htbp]
\centering
\caption{Graph-Based Digital Twin Applications for Pavement Health Monitoring and Maintenance Across Sectors}
\label{tab:graph_dt_pavement_applications}
\resizebox{\textwidth}{!}{%
\begin{tabular}{|p{3cm}|p{3cm}|p{3cm}|p{3.5cm}|p{3.5cm}|p{4.5cm}|}
\hline
\textbf{Industry Sector} & \textbf{Challenge Type} & \textbf{Digital Twin Aspect} & \textbf{Digital Twin Capability} & \textbf{Enabled Benefits} & \textbf{Real-World Outcomes} \\ \hline

Highway Transportation & Traffic-induced deterioration & Condition graph mapping & Analyze degradation under variable loads & Proactive maintenance alerts \citep{Wang2024} & Extended pavement life and minimized traffic disruption \citep{sierra2022development} \\ \hline

Airport Infrastructure & Runway surface fatigue & Sensor-integrated graph models & Monitor structural health in real time \citep{infrastructures5020018} & Safety assurance and cost-effective upkeep & Reduced delays and optimized maintenance scheduling \citep{Wang2024} \\ \hline

Port Pavements & High-impact axle loading & Material stress graph simulation & Predict surface wear based on cargo routes & Intelligent reinforcement & Fewer repairs and improved operational flow \citep{AYVAZ2021114598} \\ \hline

University Campuses & Budget constraints for maintenance & Network layout graphing & Prioritize based on usage patterns & Balanced resource allocation & Equitable upkeep and extended surface quality \citep{Zhu2024DigitalTwinPavement} \\ \hline

Theme Parks & Surface stress from crowd density & Foot-traffic heatmaps via node graphs & Predict wear patterns from pedestrian flows & Safer route design and reduced maintenance cost & Enhanced visitor experience and surface integrity \citep{s22124581} \\ \hline

Industrial Zones & Heavy vehicular stress & Load-route correlation graphs & Detect critical stress points for reinforcement & Lifecycle cost savings \citep{HUANG2009197} & Lower downtime and fewer structural failures \\ \hline

Urban Smart Cities & Multi-agency coordination gaps & Integrated infrastructure graph & Enable cross-domain planning and scheduling \citep{ma16031047} & Minimized redundancy and cost-sharing & Optimized repair timelines and citizen satisfaction \citep{su151410908} \\ \hline

Cold Region Networks & Seasonal surface cracking & Weather-data-linked graphs & Simulate freeze-thaw cycles & Preemptive repair planning & Fewer cold-induced failures and better resilience \citep{s22124581} \\ \hline

Military Bases & Strategic mobility under pressure & Resilience-ranking node graphs & Simulate disruptions and re-routing \citep{sierra2022development} & Rapid restoration capability & Ensured mission-readiness and operational continuity \\ \hline

Rural Roads & Accessibility gaps & Community-informed road graph & Prioritize underserved routes & Inclusive maintenance policies & Improved connectivity and social equity \citep{Wang2024} \\ \hline

\end{tabular}
}
\end{table}

\subsection{Sustainable Development Goals (SDGs)}

The proposed DT-GNN framework aligns closely with several United Nations Sustainable Development Goals (SDGs), highlighting its broader societal relevance beyond technical innovation. Specifically, our work contributes to SDG 9 (Industry, Innovation and Infrastructure), SDG 11 (Sustainable Cities and Communities), and SDG 13 (Climate Action).

\textbf{SDG 9} emphasizes the need for resilient infrastructure and the promotion of inclusive and sustainable industrialization. Our framework supports this goal by enabling real-time pavement condition assessment and predictive maintenance through advanced analytics. By replacing reactive repairs with data-driven, anticipatory interventions, the system enhances road longevity, minimizes infrastructure downtime, and improves resource efficiency \citep{un_sdg9}.

\textbf{SDG 11} focuses on making cities inclusive, safe, resilient, and sustainable. Urban transportation systems heavily depend on the quality and reliability of pavement infrastructure. By reducing unplanned road closures and extending pavement service life, our framework contributes to smoother urban mobility, safer transport networks, and reduced disruption for commuters and freight systems \citep{un_sdg11}. Furthermore, the integration of what-if scenario simulations allows city planners to evaluate the long-term impacts of various maintenance strategies before physical deployment, enhancing the resilience of urban infrastructure systems.

\textbf{SDG 13} calls for urgent action to combat climate change and its impacts. Poor road conditions lead to increased vehicle fuel consumption and greenhouse gas emissions due to traffic delays and inefficient routes. By optimizing maintenance timing and targeting critical pavement segments, our approach reduces unnecessary emissions and supports a more environmentally sustainable infrastructure lifecycle \citep{un_sdg13}.

Through these contributions, our DT-GNN system not only advances technical pavement monitoring capabilities but also actively supports the global agenda for sustainable development.

\subsection{ Challenges} \label{sec:Challenges-and-research-directions}
Implementing and deploying GNN-based systems in infrastructure management presents several critical challenges that guide future research directions. First, data availability and quality are significant concerns, particularly in developing countries like Bangladesh, where high-resolution, temporally consistent pavement condition data are often scarce due to limited sensor networks and historical records. Second, the computational complexity of training GNNs on large-scale networks can be prohibitive in resource-constrained environments, requiring more efficient architectures or approximation techniques. Third, model generalization remains difficult, as GNNs tend to capture localized features, making them less transferable across diverse geographical regions, pavement types, or traffic profiles; this necessitates advancements in domain adaptation and transfer learning. Fourth, implementing a real-time DT involves complex integration of IoT systems with robust data fusion mechanisms, interoperability protocols, and standardization practices, many of which are still under development in the infrastructure sector. Finally, security and privacy concerns arise when dealing with continuous data streams from infrastructure, highlighting the need for secure data handling practices, resilient system architectures, and safeguards against cyber threats and sensor malfunctions.

\subsection{Future Research Directions} 
Future research should focus on transitioning the conceptual framework into practical applications through prototyping and pilot testing. Implementing the GNN-enhanced DT in urban road sections using mobile sensors or IoT devices will help validate the model and identify data collection challenges. Additionally, developing lightweight GNN models for real-time processing, such as GraphSAGE or attention-based GNNs, will enhance scalability and deployment on edge devices in resource-limited environments. Integrating multi-modal datasets, including traffic patterns, weather conditions, and historical maintenance data, will improve the model's prediction accuracy and contextual understanding. Establishing adaptive feedback loops within the system will also enable continuous learning from real-world maintenance outcomes, further enhancing decision-making over time. Finally, assessing the policy and economic impacts of predictive maintenance strategies is crucial. Research in this area will provide insights into cost savings and social benefits, helping justify investments in such intelligent systems, particularly in developing countries like Bangladesh.

\section{Conclusion} \label{sec:conclusion}

This study presents a novel framework that integrates GNNs within a Digital Twin (DT) ecosystem to enhance pavement health monitoring and maintenance optimization. By leveraging GNNs’ ability to model complex graph-structured dependencies, the proposed approach significantly improves the predictive performance of pavement distress levels compared to traditional regression models. The experimental results demonstrated that the GNN model achieved the highest R² score of 0.3798, outperforming baseline models including Random Forest and Gradient Boosting. The model also attained a balanced mean absolute error (MAE) of 31.34 and root mean square error (RMSE) of 38.93, indicating a strong capability to generalize under complex data conditions. Nevertheless, the model showed limitations in classifying moderate distress levels accurately and exhibited occasional large prediction errors, underscoring the need for further refinement in handling ambiguous or underrepresented data regions. By enabling more precise real-time analysis and predictive forecasting, this integrated GNN-DT framework offers a promising pathway to optimize maintenance strategies and resource allocation, ultimately enhancing infrastructure longevity and public safety. Although challenges related to data sparsity and computational demands remain, the results confirm the feasibility and advantages of incorporating graph-based deep learning methods into pavement management systems. Future research should aim to improve classification robustness, expand data diversity, and develop scalable implementations suitable for operational deployment. With ongoing advances in sensing technologies, edge computing, and machine learning, the adoption of GNN-enhanced Digital Twins for smart pavement infrastructure management is increasingly attainable.
\clearpage
\bibliography{main}

\appendix
\appendix
\end{document}